\documentclass[lettersize,journal]{IEEEtran}
\usepackage{amsmath,amsfonts}
\usepackage{algorithmic}
\usepackage{hyperref}
\usepackage{algorithm}
\usepackage{array}
\usepackage[caption=false,font=normalsize,labelfont=sf,textfont=sf]{subfig}
\usepackage{textcomp}
\usepackage{stfloats}
\usepackage{url}
\usepackage{verbatim}
\usepackage{graphicx}
\usepackage{cite}
\usepackage{subfig}
\usepackage[
singlelinecheck=false % <-- important
]{caption}
\usepackage{etoolbox}
\makeatletter
\patchcmd{\@makecaption}
  {\scshape}
  {}
  {}
  {}
\makeatother

\DeclareRobustCommand{\IEEEauthorrefmark}[1]{\smash{\textsuperscript{\footnotesize #1}}}

\begin{document}
\bstctlcite{IEEEexample:BSTcontrol}

\title{Dynamic Neural Curiosity Enhances Learning Flexibility for Autonomous Goal Discovery}

%\author{Quentin Houbre and Roel Pieters$^{1}$% <-this % stops a space

\author{Quentin Houbre\IEEEauthorrefmark{1} and Roel Pieters\IEEEauthorrefmark{1}

\IEEEauthorblockA{%
    \IEEEauthorrefmark{1}Tampere University, Tampere, Finland.
  }
\thanks{$^{1}$Department of Automation Technology and Mechanical Engineering, Cognitive Robotics, Tampere University, 33720, Tampere, Finland; {\tt\small firstname.surname@tuni.fi}}
}

% The paper headers
%\markboth{Journal of \LaTeX\ Class Files,~Vol.~14, No.~8, August~2021}%
%{Shell \MakeLowercase{\textit{et al.}}: A Sample Article Using IEEEtran.cls for IEEE Journals}

%\IEEEpubid{0000--0000/00\$00.00~\copyright~2021 IEEE}
% Remember, if you use this you must call \IEEEpubidadjcol in the second
% column for its text to clear the IEEEpubid mark.

\maketitle

\begin{abstract}
The autonomous learning of new goals in robotics remains a complex issue to address. Here, we propose a model where curiosity influences learning flexibility. To do so, this paper proposes to root curiosity and attention together, taking inspiration from the Locus Coeruleus-Norepinephrine system along with various cognitive processes such as cognitive persistence and visual habituation. We apply our approach by experimenting with a simulated robotic arm on a set of objects with varying difficulty. The robot first discovers new goals via bottom-up attention through motor babbling with an inhibition of return mechanism, then engages in the learning of goals due to neural activity arising within the curiosity mechanism. The architecture is modeled with dynamic neural fields, and the learning of goals such as pushing objects in diverse directions is supported by the use of forward and inverse models implemented by multilayer perceptrons. The adoption of dynamic neural fields to model curiosity, habituation, and persistence allows the robot to demonstrate various learning trajectories depending on the object. In addition, the approach exhibits interesting properties with respect to the learning of similar goals and the continuous switch between exploration and exploitation.
\end{abstract}

\begin{IEEEkeywords}
Intrinsic motivation, curiosity, attention, task engagement, autonomous goal learning, dynamic neural fields. 
\end{IEEEkeywords}

\section{Introduction}

Developmental robotics takes inspiration from various fields such as developmental psychology, neuroscience, machine learning, or philosophy. A popular domain of interest in the field and one of the core ideas of this study is called \textit{Curiosity}. Curiosity can be seen as a particular case of intrinsic motivation (IM), and researchers have demonstrated that IM could be an efficient drive for the autonomous learning of behavior \cite{oudeyer_intrinsic_2007}. Moreover, the acquisition of different behaviors follows a developmental trajectory \cite{oudeyer2010intrinsically, oudeyer2013intrinsically}. Nowadays, there are multiple ways of modeling curiosity, and one of them consists of training predictors, such as forward and inverse models to create learning progress based on the prediction error. The use of curiosity in robotics supports the generation and learning of goals that increase in complexity \cite{baranes2010,BARANES2013}.
Goals are typically selected and managed via a top-down approach, where a central system monitors and determines the selection and learning time of the goals. Despite demonstrating impressive results, this approach narrows the understanding of curiosity, as it involves the interaction between various cognitive processes.

Along with curiosity, attention is a fundamental cognitive process. In fact, attention allows humans to focus on specific elements of our environment. However, attention is a broad term that gathers several specific mechanisms. For example, visual attention is the ability to maintain gaze on prominent stimuli and develops during the first months after birth \cite{richards1985development}, \cite{atkinson1992changes}. To help search for a specific visual stimulus, \textit{inhibition of return} is a well-studied mechanism that naturally prevents a person from looking twice at the same location in a short time interval \cite{posner_inhibition_1985}. In addition, we can distinguish two types of visual attention: exogenous, driven by external stimuli, or endogenous and goal-driven \cite{posner1971components}, \cite{maclean2009interactions}. These two types of attention are often referred to as bottom-up for the former and top-down for the latter, and inhibition of return can occur during both, but whether there are shared neural pathways between them to occur is still subject to debate \cite{chica_effects_2009}. In the brain, arousal is mediated by the Locus Coeruleus-Norepinephrine system (LC-NE) and demonstrates different patterns of neural activity. According to adaptive gain theory \cite{aston-jones_integrative_2005}, LC neurons exhibit two modes of activation: tonic and phasic. During tonic activation, the subject disengages from the current task and engages in an exploratory behavior. In contrast, phasic LC activation engages in exploitation and thus is driven by task-specific outcomes. IM has been shown to be involved in the tonic and phasic activation of dopamine \cite{gottlieb2016motivated}, overlapping certain neural pathways with noradrenaline (norepinephrine) \cite{ranjbar2020dopamine} and thus could interact directly with the LC-NE system. If curiosity and more especially its learning progress component is related to attention, then the LC-NE system is a potential starting point.

Here, we propose a robotic cognitive architecture for the online discovery and learning of goals, where a robotic arm is learning how to interact with simple objects. To do so, we identify and model several cognitive mechanisms to determine if the robot should explore and discover potential goals through bottom-up attention, or exploit and learn to achieve these goals. We introduce a component inspired by the LC-NE system to dynamically switch between goal discovery and learning. During both exploration and exploitation, an inhibition of return mechanism generates a set of actions toward the object location until producing a change in the environment. If the inhibition of return is shared by exploration and exploitation, there is, however, a conceptual difference between them since the discovery of a new goal in the environment rests on exogenous attention, and on a top-down and goal-driven approach for the learning of these goals. For the exploration stage, we complete the attentional mechanism connected to the LC-NE system with a habituation paradigm \cite{sophian_habituation_1980}, where the discovery of a new goal produces a tonic activation and limits the exploration time. To balance this process, the absence of a new goal signals a shift from tonic to phasic activation. During the latter, the learning of a goal occurs with the use of simple neural networks that implement a forward and inverse model. The output of these neural networks determines an error associated with a goal and its learning progress. Except for the forward and inverse models, the entire architecture is designed with dynamic neural fields \cite{schoner2016dynamic}. The goals are then represented within the neural fields by the neural activation of their learning progress and are projected to the LC-NE system, which then chooses to exploit the goal with the highest learning progress. By design, the model follows an enactive approach by continuously pushing the robot to interact with its environment, thus shaping its cognition using several forms of working memory. We will demonstrate that when similar goals are discovered, the robot autonomously only focuses on the one with the highest error and slowly forgets about the other. We evaluated several neural parameters such as persistence, prediction error inhibition, and habituation that modulate and optimize the learning of goals, depending on the objects. To our knowledge, this is the first attempt to model intrinsic motivation with dynamic neural fields and thus support the robot to express different learning behaviors by tuning intrinsic properties of the dynamic neural fields.

This paper will provide a theoretical background according to the literature in Section \ref{related} before presenting the architecture of the experiments (Section \ref{arch}). Then, we will present the experimental setup with the frameworks used in Section \ref{exp} followed by the different results (Section \ref{res}). Finally, we will discuss the advantages and limitations of the architecture (Section \ref{disc}) before concluding.

\section{Related work}\label{related}
\subsection{Curiosity and Intrinsic motivation}
In terms of biological perspective, intrinsic motivation (IM) should be distinguished from extrinsic motivation \cite{baldassarre_what_2011}. The former can be seen as a way for the brain to monitor the learning efficiency of a particular knowledge, while the latter directly guides the learning itself through external cues/rewards. Then, there are different metrics to compute intrinsic motivation \cite{oudeyertypo}, separating knowledge-based systems from competence-based systems \cite{mirolli_functions_2013}. In the literature, there is a consensus that classifies curiosity as a particular case of intrinsic motivation \cite{oudeyer2016intrinsic}. In fact, curiosity focuses on the learning progress hypothesis  \cite{schmidhuber1991curious}, \cite{oudeyer_intrinsic_2007}, and formulates it as the evolution of prediction error while learning a task. In practice, curiosity can be used to learn a world model by focusing on the sensorimotor experience \cite{lefort_active_2015}. The terms goal and skill often refer to the same notion, even if a skill can be formed by the composition of several goals. For example, the grasping skill for a robot arm is formed by several goals, such as reaching, opening the gripper, adjusting the end-effector position, and closing the gripper. GRAIL \cite{santucci_grail_2016} is a cognitive architecture operating at different levels where the IM used is competency-based, determining the best goal to pursue depending on how well the learning occurs. Intrinsic motivation in robotics is also used to form hypotheses in developmental psychology. For example, a developmental robotics model identified IM to explain the sudden improvement of tool-use in children \cite{seepanomwan_intrinsic_2020}. However, the discovery of goals in visually rich environments is problematic as the exploration space rapidly expands, and an intrinsically motivated system loses efficiency. To address this issue, researchers are applying a set of dimension reduction methods. One of the first studies to overcome these limitations is the case of a robot arm interacting with a ball \cite{laversanne-finot_intrinsically_2021}. In this case, a variational autoencoder is trained, and the goals are sampled from the resulting goal space. Other works in the domain and related to robot grasping extended this concept by linking and compressing the object features, the action, and the sensori outcomes together \cite{sener_exploration_2021}. By doing so, the robot creates a latent space where the learning of forward and inverse models is facilitated to achieve grasping. Working with high-dimensional spaces and intrinsic motivation is far from trivial, and the sampling problem can gain insight from a more developmental point of view. In these studies, it is necessary to perform several rounds of exploration to build up the different learning progress of goals and then select them for learning. However, it is not clear how the switch between discovery and learning takes place in childhood. In our work, we propose that the prediction errors contribute to some extent to easing this shift. Intrinsic motivation is a reproducing force of developmental patterns \cite{oudeyer_how_2016}, but actively and efficiently sampling the environment remains difficult to tackle \cite{gottlieb_towards_2018}. Finally, the top-down approach to select and control the learning of goals creates the \textit{homunculus} problem \cite{verbruggen}. In cognitive science, this principle is known to limit the understanding of a phenomenon, especially how the cognitive control of goals might be achieved. To address this issue, our research work search to model curiosity as a set of interconnected executive functions \cite{logan_executive_1985,monsell_control_2000} to provide flexibility for the discovery and learning of goals, and to apprehend the underlying question of cognitive control. As a starting point, researchers demonstrated that attention is involved in IM, and provides a perspective on a possible neural basis of intrinsic motivation \cite{di_domenico_emerging_2017}. 

\subsection{Attention and inhibition of return}

Visual attention is essential for an individual to focus on an unexpected event and to optimize cognitive processes on a specific task \cite{maclean2009interactions}. This distinction separates exogenous (bottom-up) from endogenous (top-down) attention. If motivation can sharpen exogenous spatial attention under certain conditions \cite{engelmann_motivation_2014}, the modulation of top-down attention can enhance the learning of a particular task \cite{wulf_simply_2003, wulf_directing_2001, poolton_benefits_2006}. The neural basis for these distinct processes is, for a significant part, shared \cite{peelen_endogenous_2004}, but not entirely \cite{chica_two_2013}. Regarding the coupling between attention and intrinsic motivation, a study used reinforcement learning to allocate and shift attention \cite{di2014role}. Closer to our approach, researchers used bottom-up attention to learn object affordances and decompose a task into sub-goals \cite{baldassarre2019embodied}. Here, we choose to focus on the inhibition of return effect (IOR) to facilitate attention toward an object and ease the discovery of new goals. In the literature, the general mechanism of exploration is often summarized as random and direct exploration \cite{wilson2020deep,wilson2021balancing} where the former consists of producing random actions with a high uncertainty but potentially large expected reward. The latter directs exploration toward more certain rewards but necessitates prior information. With curiosity in robotics, this difference can be translated to a motor babbling behavior for random exploration and the building of representations for direct exploration. For example, a robot can learn a representation of a diverse set of goals encountered during random exploration to generate new goals \cite{sener_exploration_2021}. Here, our purpose is to propose an additional exploration method that can reduce the uncertainty of random exploration via the inhibition of return effect. This mechanism was pointed out by Posner \cite{posner_inhibition_1985} in visual attention when he discovered that the return of attention to a cue previously attended expresses a longer time. On the contrary, there is a facilitation (faster response time) if the time interval between cue-targets is short. This consequence endows the brain with a natural foraging mechanism of visual stimuli \cite{tipper_object-centred_1991}, but it remains necessary to proceed to an attentional disengagement to observe the IOR effect  \cite{klein_inhibition_2000}. However, it appears that an endogenous disengagement of attention is not enough in order to observe the effect and is instead always relying on exogenous disengagement of attention \cite{lupianez2013inhibition}. In addition, it is possible to induce an inhibitory of return for both exogenous (uncued location) \cite{chica_dissociating_2006}, \cite{chica_effects_2009} and endogenous (cued location) attention, but the effect seems highly correlated to the cue's saliency and thus relies on an important exogenous component \cite{henderickx_involvement_2012}. Regarding IOR in action production, there is no consensus on whether the effect is directly or partially involved. Several research reproduced the reaction time of IOR during a hand reaching movement \cite{tipper1998action} and even showed that reaching paths were biased toward the cue locations under certain conditions \cite{howard1999inhibition}. Directly connected to our research, it has been determined that IOR effects can be applied to objects \cite{grison2004object}. Even more, inhibition can interact with working memory to mediate goal-directed action toward an object, optimizing performance for future interaction. In consequence, we intend to generate and use IOR around the object location so the robot can generate goal-driven actions. To be able to identify these two forms of attention and decide to engage in exploration, we take inspiration from the Locus Coeruleus.

\subsection{Exploration and exploitation}

The locus coeruleus-norepinephrine system (LC-NE) is a brainstem nucleus that is widely connected to the neocortex. The first unifying approach with respect to its role and functions is the adaptive gain theory \cite{aston-jones_integrative_2005}. The theory demonstrates two different neural activations, depending on the arousal state and the relation with task engagement. More precisely, the LC-NE system exhibits a tonic neural activation during exploration, raising the level of arousal and thus bottom-up attention to salient events \cite{sara_orienting_2012}. Here, the exploration strategy consists of generating goal poses around the object with IOR and observing the possible outcomes.  Phasic activation suggests a goal-oriented behavior (exploitation) with a focus on the task (top-down attention). Then, the LC-NE system is a crossroad between attentional processes and decision making. Aston-Jones and Cohen determine the emergence of a tonic activation when the utility of a task vanishes. The LC-NE system receives direct connection from the anterior cingulate (Acc) and the orbitofrontal cortices (OFC), which are directly involved in task-related utility and decision making \cite{wallis_contrasting_2011}, \cite{khani_partially_2014}. More specifically, the ACC appears to represent predictions of value about reward and, more generally, uncertainty \cite{monosov_anterior_2017}. There is a consensus on the recognition of the basal ganglia as the reward processing site that can support reinforcement learning \cite{bar2003information}. Several robotics experiments take inspiration from the phasic delivery of dopamine to signal rewards based on prediction errors \cite{santucci2010biological}, \cite{mirolli2013phasic}. We can observe a similarity between the phasic delivery of noradrenaline from the LC and the phasic dopamine-based signal emerging from the basal ganglia, especially after pointing out that dopamine and noradrenaline share neural pathways \cite{ranjbar-slamloo_dopamine_2020}. In addition, researchers demonstrate that if dopamine and noradrenaline are related to reward prediction, only the latter is predictive of task engagement and is strongly activated when cues indicate a new task condition. This means that, contrary to dopamine, noradrenaline is strongly involved in motivating current or future engagement. The Acc directly projects prediction information about the task utility to the LC-NE system, which in turn decides if the utility is high enough to drive the neural activation in a phasic mode, hence to an exploitatory stage. Researchers proposed an alternative view of the LC by describing its phasic activation as a reset network capable of reorienting resources to a new task \cite{BOURET2005,sara_orienting_2012} while still being strongly sensitive to rewards \cite{bouret_sensitivity_2015,jahn_noradrenergic_2020}. Thus, integrating the Aston-Jones LC system theory with prediction-error sensitivity appears promising for enhancing curiosity-driven robotics. It will allow the robot to engage in exploitation depending on uncertainty, meaning on how accurate the robot's predictions are, based on the learning progress. As a consequence, the LC model will follow a phasic neural activation, thus autonomously tuning attention to modulate the level of arousal and engage in the learning of a goal. Recent neuroscience research highlights the role of the median raphe nucleus (MRN), a brainstem nucleus, in perseverative, exploratory, and disengaged states \cite{ahmadlou_subcortical_2025}. In their study, they discovered that GABAergic and serotonergic neurons influence perseverative states and exploration, respectively. There are currently no studies that investigate the influence of the MRN on the LC-NE system and vice versa. Their close proximity in the brain and their implication in exploratory and exploitative states could indicate a relationship, but this hypothesis is currently a speculation. We intend to use these findings to modulate exploration and regulate goal learning by investigating the role of excitatory and inhibitory neurons, respectively. To do so, we first identify habituation as a cognitive mechanism that can regulate exploration. Then, we investigate the process of persistence (i.e., perseverance) to regulate the allocated time for the learning of a goal.

\subsection{Habituation and Persistence}

In order to limit the time spent on exploration, our approach introduces a cognitive process called habituation. In fact, there is yet no evidence on how the MRN might influence the LC to switch from a tonic mode with high arousal to a phasic mode (exploitation). Here, we propose to model this switch by a habituation paradigm where the phasic activation from LC happens only when the robot's perception of an object does not produce a novelty effect anymore. Researchers, particularly in developmental psychology, study habituation by measuring the duration of infant attention to novel stimuli \cite{pecheux_habituation_1983}. The research on the topic first focused on the link between perception familiarity and exploratory behavior \cite{hunter1983effects} and concluded that infants who have become familiar with some objects prefer to look at new ones. A similar study \cite{ruff1986components} confirms that babies between the ages of seven and twelve months have a decrease in attention when objects become familiar. Interestingly, the study determined that the look duration did not depend on age but rather on the familiarity of the object. Habituation is an essential mechanism for cognitive function \cite{schmid_habituation_2015}, and simulation of this mechanism supports the reproduction of developmental studies \cite{schoner2006using}. In this work, we will adopt and modify the model proposed by Perone and Spencer \cite{perone2013autonomy} to avoid the dishabituation of objects when they are out of sight. In addition, the proposed model will avoid reproducing the familiarity effect for this study, as it can complicate the study of the goal discovery and learning stages.

Cognitive persistence is a fundamental aptitude if one wants to learn a new goal. Observing how individuals persevere to learn something new can help to decide how and whether learning is actually fruitful. Regarding developmental psychology, the measure of persistence is investigated by assigning tasks to infants and evaluating how long they engage in a learning activity \cite{yarrow_infants_1982}. Here, researchers established that there is a link between an infant's persistence and competence. This means that infants who were more stimulated at home persist more in goal-oriented actions. In contrast, infants with cognitive delay demonstrated less persistence, as well as a desire to participate in less challenging learning activities \cite{hupp1991persistence}. From a neuroscience perspective, Teubner-Rhodes investigated this ability to overcome task difficulty in adults \cite{teubner-rhodes_cognitive_2017}. Using fMRI, she located persistence in the prefrontal cortex and concluded that adults with high perseverance demonstrated better performance and adaptive control for the task. Later, a study focused on persistence for multilingual people \cite{teubner-rhodes_cognitive_2020} and located this process in the inferior frontal gyrus and dorsal anterior cingulate cortex (dACC). This could suggest that, in addition to dealing with uncertainty and prediction error, ACC is also involved in cognitive persistence. The MRN is involved in cognitive persistence \cite{ahmadlou_subcortical_2025}, and the suppression of inhibitory neurons leads to more perseverative behavior. More specifically, the serotonin neurotransmitter appears to have a strong influence on perseverative states through widespread inhibition \cite{matias2017activity,lottem2018activation}. These studies question the roles of the MRN along with the LC as they seem to partially overlap regarding task engagement, but also seem complementary during learning. 

In this work, we intend to propose a simple model of persistence (i.e., perseverance) to regulate the learning of a goal via the involvement of prediction errors. The purpose is to endow the robot with the ability to dynamically allocate time to learn a goal depending on learning progress and prediction error. Each aspect of cognition described in this section will be modeled with dynamic field theory.

\subsection{Dynamic Field Theory}

In this work, we model the different cognitive processes with neural dynamics and, more especially, Dynamic Field Theory (DFT) \cite{schoner2016dynamic}. By doing so, the system exhibits a range of different behaviors by tuning only the intrinsic parameters of the neural fields. The theoretical framework mathematically models the evolution and activity in time of large populations of neurons. The approach has been successful in modeling complex cognitive functions, such as visual working memory \cite{BussMagnottaPennyEtAl2021}, intentionality \cite{TekulveSchoner2020}, or motor habituation \cite{AerdkerFengSchoner2022}. The formulation and implementation of DFT are provided in the Appendix \ref{app:dft}. Regarding the modeling of motions, we adopted the dynamic movement primitive framework.

\subsection{Dynamic Movement Primitive}

In this work, we control the motion of a robotic arm using the Dynamic Movement Primitive Framework (DMP) \cite{ijspeert_dynamical_2013}, as it can exhibit a wide range of motions with a limited number of parameters \cite{pastor2013dynamic}. The approach uses a set of differential equations to represent a movement, with the advantage of being robust against perturbation. More precisely, we adopt the formalism introduced by Pastor \cite{pastor2009learning}, which allows the generalization of a motion by adapting a \textit{start} and \textit{goal} parameter. More details about the method and its implementation can be found in the Appendix \ref{app:dmp}.

\section{Architecture}\label{arch}
\subsection{Overview}

%\begin{figure}[h]
%\centering
%\includegraphics[width=0.45\textwidth]{archi_uni.png}
%\caption{General Architecture\label{fig:gen}}
%\end{figure}

\begin{figure}[ht]
\centering
\includegraphics[width=0.48\textwidth]{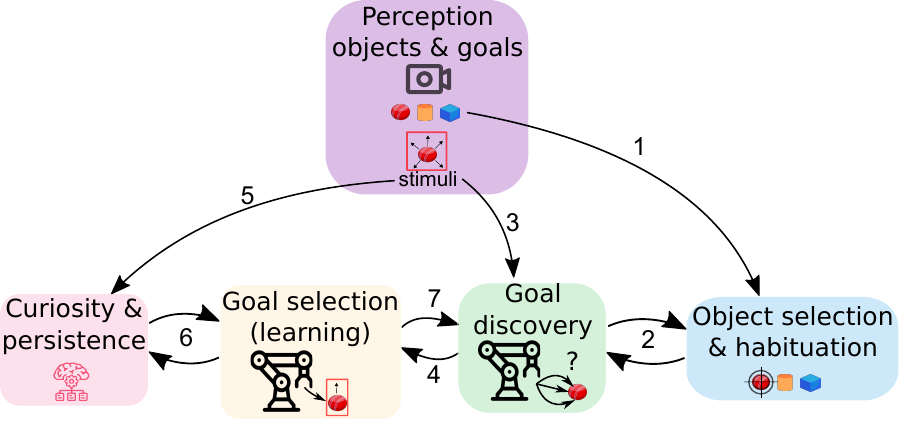}
\caption{Timecourse for the discovery and learning of goals.\label{fig:steps}}
\end{figure}

The Locus Coeruleus (LC) architecture is an essential component of our model. In the brain, this nucleus is directly responsible for attention and task engagement. Here, we link intrinsic motivation and LC activity through engagement in learning new goals with different types of attention.  In this section, we propose to articulate and describe how these cognitive processes might work together.

\begin{figure*}
\centering
\includegraphics[width=0.95\textwidth]{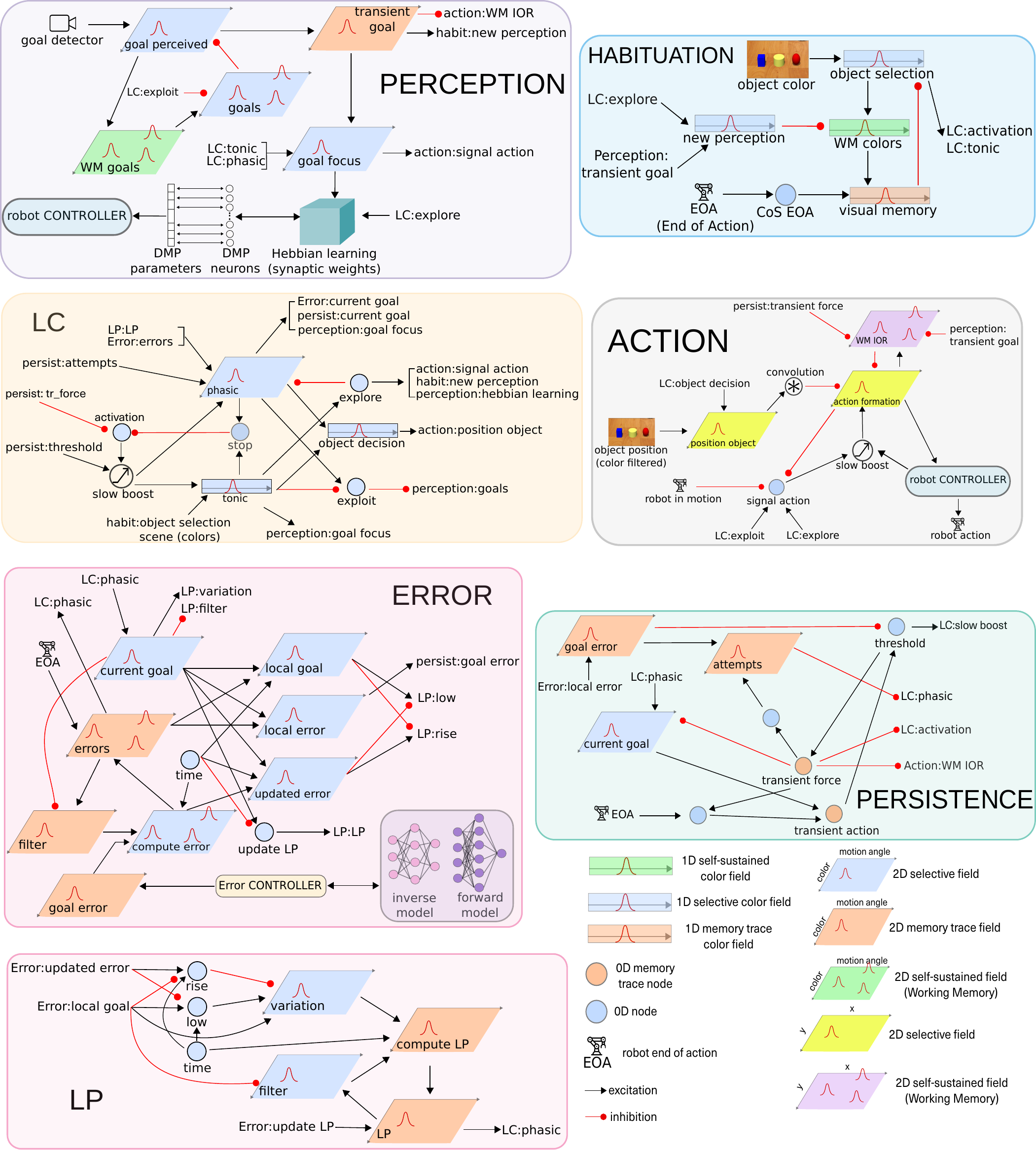}
\caption{Overview of the general architecture with the cognitive processes. Connections between the mechanisms are detailed within each component.\label{fig:general}}
\end{figure*}

The LC exhibits two modes of neural activation: tonic for exogenous attention and phasic spiking mode for top-down attention during task engagement. Due to the time-scale abstraction level, the DFT framework cannot precisely differentiate them as well as biological neurons. Then, we define a tonic activation as a sustained and uninterrupted excitation activity, while the phasic mode exhibits a more sparse and localized activation over time. Figure \ref{fig:steps} depicts the architecture steps from goal discovery with bottom-up attention to learning with curiosity. At first, the robot detects objects with the camera and applies a color filter. This information is formalized as neural activity and projected to the habituation component that selects an object to begin exploration. Then, the goal discovery process begins by performing action babbling around the object. During this stage, we can consider that the type of attention is exogenous, since the robot is sensitive to any stimuli related to the selected object. At the same time, a habituation mechanism reduces the sensitivity towards the objects, thus reducing the neural activity within the LC. The robot then selects a goal to learn with the help of the curiosity and persistence mechanisms. At that step, there is a shift from bottom-up to top-down attention, since the model only focuses on observing outcomes from a particular object. Finally, the robot can switch attention to a new object and start over the process of goal discovery. However, this last step could be ignored, and the robot will choose to discover new goals on a previously seen object. 

\subsection{Locus Coeruleus}\label{sec:lc}

The LC mechanism coordinates the engagement in the discovery or learning of goals. At any time, the active node always exhibits a suprathreshold level of activation and excites the slow boost component (see Fig. \ref{fig:general}), a particular form of memory trace that evolves depending on two node inputs (Appendix \ref{app:dft}). This memory trace builds activation when the active node projects excitation, maintains the current level if both connected nodes are down, and decays the activation if only the threshold node is up. At first, the slow boost module slowly increases the resting level of the phasic and tonic neural field (NF) and stops when a peak of activation appears, triggering the stop node. The tonic activity of neurons is depicted with the tonic NF and the excitation incoming from the habituation mechanism (object selection NF). This neural field exhibits a peak of activation as long as the robot is not accustomed to an object. The phasic NF gathers activations from the error and learning progress modules, but with a higher resting level. This means that after discovering goals, the tonic NF will be the first to see the emergence of suprathreshold activations. Tonic NF has a resting level of -2 and phasic NF -1.05. This implies that LC can see a peak in tonic NF when cumulative learning progress and error are below 0.05. This indicates that the LC can go back to exploration when all the goals have been learned. After discovering several goals through bottom-up attention and being habituated to an object, the robot does not show any learning progress. Then, it is necessary to bind the error memory trace to the phasic NF to bootstrap the learning of previously discovered goals. We empirically chose the gain factor so that the robot can engage in learning a goal if there is no learning progress ($\approx$ 30 percents of the errors present in the error module). However, we keep the error contribution low enough to favor the learning of a task that has a small learning progress. By doing so, the robot can begin to learn goals that do not exhibit any learning progress, instead of performing several stages of exploitation for each goal beforehand to build up the different learning progresses. The one-dimensional object decision NF represents the object's color in which the robot should explore or engage in learning. 

Here, we prioritize exploration in the same manner that exogenous attention seems to be operating within the LC. As mentioned in the literature (\ref{related}), the engagement in a task can only operate if there are no salient stimuli and thus a low level of bottom-up attention. The priority of exploration comes with the tonic activation. Our approach demonstrates sustained activation (i.e., tonic) while discovering new goals, thus preventing the possibility to learn a specific goal. During exploitation, the model selects a goal to learn regularly. This process stops any activation within the phasic NF and raises again the resting level to select the goal with the highest learning progress (or error in the absence of any learning progress). In this model, the persistence mechanism directly influences the selection of the goal and depicts the phasic activity. Finally, if the robot is habituated and has learned all the discovered goals, the tonic NF will select a known object to conduct the exploration again (Appendix \ref{app:habituation}).

\subsection{Perception and habituation}\label{sec:habit}

The perception module processes the color and motion of the object. The colors of objects in sight are formalized as inputs to a one-dimensional field and sent to the habituation mechanism, where neural dynamics can perform the selection of an object. A goal is formalized by the motion direction of an object. This motion is characterized by an angle between a reference vector and a vector representing the motion of the object along the dimensions $x$ and $y$ (see Appendix \ref{app:perception}). Here, we voluntarily design the goal as simply as possible to study the underlying mechanism of goal discovery and learning. The goal space is then formalized by a 2D neural field, where the vertical and horizontal dimensions represent the object color and motion angle as a goal, respectively (Figure \ref{fig:general}). It is important to emphasize that the detection of the object's motion takes place in real time (i.e., while the robot performs the action), delivering a sustained input to the goal perceived NF (Neural Field). In Figure \ref{fig:general}, an inhibition of return occurs with the goal perceived NF, the WM goals, and the NF goals to avoid perceiving stimuli as new when the goal has already been discovered. During exploration, and if the stimulus is new, the robot controller generates a dynamic motion primitive (DMP) and activates the associated DMP neuron through a one-to-one connection. The Hebbian learning model links the created DMP to the observed outcome (goal focus NF) using a reward-gated rule similar to \cite{TekulveSchoner2020}:
\begin{equation}
\begin{split}
    \dot{w}_{{dmp}_i,{col,ang}} = -\eta \, \sigma(LC:explore) \, \sigma(dmp_{i}) \\ ({w}_{{dmp}_i,{col,ang}} - \sigma(g(col,ang)))
\end{split}
\end{equation}
Learning occurs when there are activations from the DMP neuron, the explore node, and the goal focus NF. They are respectively denoted by \textit{$\sigma(dmp_i)$}, \textit{$\sigma(lc:explore)$} and \textit{$\sigma(g(col,ang))$}. The term \textit{$\sigma(g(col,ang))$} represents the sigmoid activation of the goal focus NF at the object color (\textit{$col$}) and angle (\textit{$ang$}) location. When the robot engages in learning a goal (i.e., exploitation), the goals NF is inhibited, so a peak of activation can emerge within the goal focus neural field. The activation then spreads through the synaptic weights and triggers the DMP neuron corresponding to a particular goal. 

The habituation mechanism is closely related to the perception module and is inspired by the work of Perone and Spencer \cite{perone2013autonomy}. The mechanism slowly habituates the robot to repeated stimuli, such as observing the same goal from the same object. The LC switches from a tonic mode to a phasic mode because of habituation. The color of the object serves as a feature within the object selection NF (Fig. \ref{fig:general}). The new perception field receives activation from the goal color dimension, as well as from the explore node inside the LC module, supporting a suprathreshold activation only during exploration. The Condition of Satisfaction Node (CoS) is only active for a brief period of time when the robot has finished an action (End of Action). This node corresponds to the term \textit{a(t)} in Eqn. \eqref{eq:mt}. When a new goal occurs in the scene, visual memory slowly decays at the color location, reducing the inhibition of the object selection field. In contrast, if an action results without the perception of a new goal, visual memory builds up an activation that will inhibit object selection NF. In practice, this implies that the robot keeps the exploration of new goals as long as the inhibition is weak enough within the object selection NF. The main difference from the habituation model of Perone and Spencer rests in the coupling between the WM colors and the visual memory trace. Here, the inhibition of the visual memory trace is directly linked to the WM colors, while their model associates an inhibitory layer with a Hebbian mechanism to the object selection NF. In our model, it results in the object remaining habituated even if it is out of sight, and with the possibility of decaying inhibition when new stimuli are observed. 

Here, the use of dynamic neural fields provides the possibility of designing a more robust perception system. The inhibition of return prevents the robot from identifying a previously discovered goal as a new one if it is seen again during exploration. By tuning the neural field parameters (i.e., the standard deviation of peaks), it is possible to adjust the likelihood of two goals. With an elevated standard deviation, two similar goals have a lower probability of being distinguished from each other. In addition, dynamic field theory demonstrated excellent results in modeling brain processes as habituation. This is, to our knowledge, the first curiosity-driven architecture that integrates such a mechanism to favor exploration toward novel objects. The following section will describe the action and the persistence system.

\subsection{Action and Persistence}

The role of this process is to perform action babbling around the object located in the scene (Fig. \ref{fig:general}). During exploration, the module generates two poses ($x_1$,$y_1$) and ($x_{goal}$,$y_{goal}$) through which the end effector will go from a resting position ($x_{start}$,$y_{start}$). In this work, the \textit{z} dimension is fixed at one centimeter above the table for a generated pose. The robot records the end effector position from the resting pose until reaching the second pose and creates a DMP if it leads to the discovery of a new stimulus. During goal learning, the robot activates the DMP parameters of a goal and only needs to generate a single pose. 

\begin{figure}[ht]
\centering
\subfloat{\includegraphics[width=1.1in]{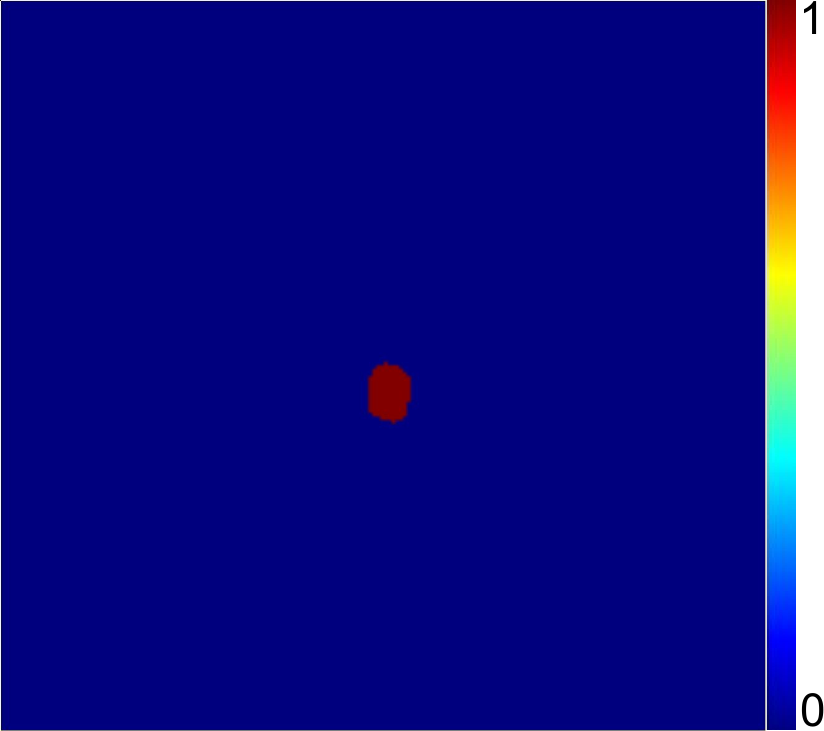}%
\label{fig:inh_conv}}
\hfil
\subfloat{\includegraphics[width=1.18in]{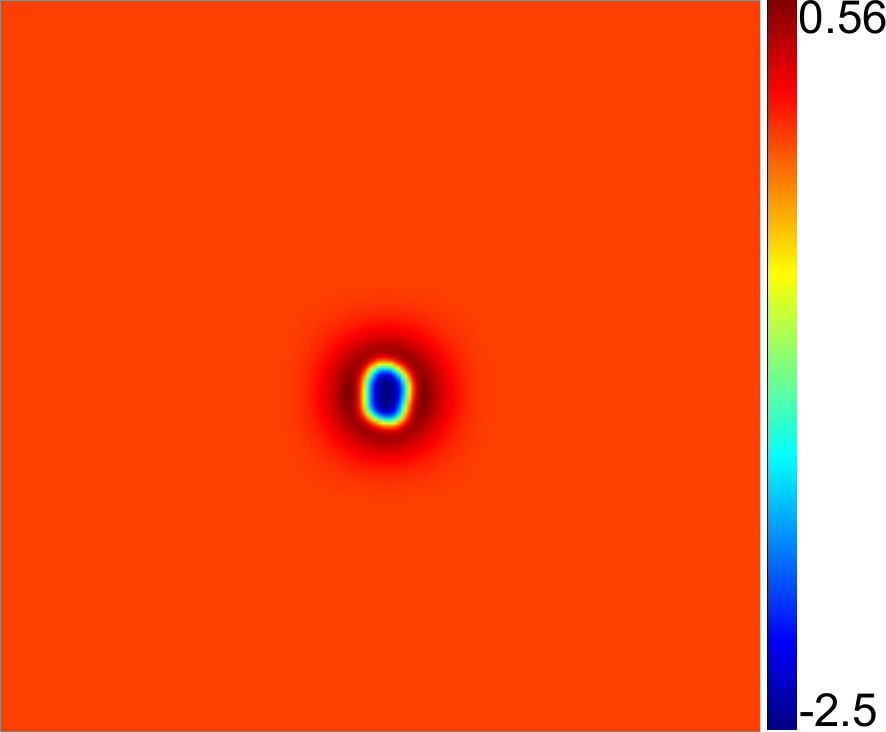}%
\label{fig:peak_action}}
\hfil
\subfloat{\includegraphics[width=1.1in]{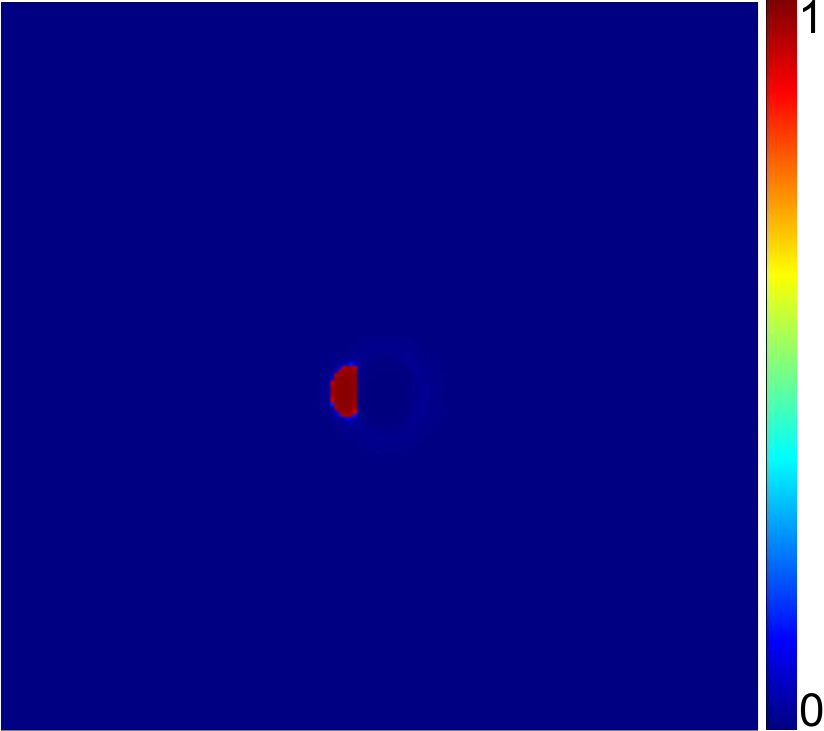}}%
\label{fig:conv_final}
\caption{(Left) Object input (cube) to position object NF in the action component. (Middle) Mexican-hat convolution with inhibition applied to the object (input to action formation NF). (Right) Generated pose after raising the resting level within the action formation NF. \label{fig:action_nf}}

\end{figure}

Regarding object perception, a filter based on the selected object is applied (i.e., from the habituation component). After a transformation to align the image to the robot space coordinate system, the filtered image serves as input to the position object NF (Fig. \ref{fig:action_nf}-left). Then, we apply a specific convolution between the position object output and the action formation NF. The result of this convolution then serves as an inhibition to action formation NF for the action babbling behavior (see Fig. \ref{fig:action_nf}-center). 

To generate a pose, we use a \textit{slow boost} component as seen in the LC module (see Section \ref{sec:lc}). The slow boost builds up an activation if signal action is active, and the activation decays when the robot signals the end of an action (robot controller). The reversed Mexican-hat shape input to action formation favors activation peaks around the location of the object (Fig. \ref{fig:action_nf}-right). The signal action node receives inhibition when the robot is moving and when a peak emerges within the action formation NF. The working memory keeps track of the activations in time and projects the inhibition back to action formation. Together, these two fields provide an inhibition of return effect, avoiding selecting the same pose twice. Here, we provide a specific value for the global inhibition of working memory (Appendix \ref{app:dft}), so several suprathreshold activations can take place at the same time without overpopulation. This means that only a limited number of activations can be self-sustained. If a new peak emerges, one of the "old" activations dissipates. The robot controller is in charge of delivering excitation after two poses during exploration and after a single pose for exploitation. Working memory is inhibited in the event of new stimuli (transient goal) and when the persistence mechanism reaches the threshold.

The persistence module carries a rhythm to the action module by signaling when to reset the inhibition of return while learning a specific goal. In the case where the goal error does not decrease, this module stops any further trial to learn a particular goal and clear the memory of all the generated poses. The core of the process lies with the threshold, force, and action nodes. The threshold node is a classic zero-dimensional neural field, transient action and force nodes are memory traces with a slow build-up and fast decay for the former, and a quick build-up and long decay for the latter. The threshold node receives inhibition from the local error level (i.e., the error of the current goal) and excitation from the transient action. However, the current goal NF output is either zero or one because of the use of the absolute sigmoid as kernel activation. The output of the current goal is always higher than the local goal error that is bounded by the hyperbolic tangent function (Section \ref{im}). After each movement, the level of activation within the transient action gradually increases and spreads to the threshold node. Once the level of transient action overcomes the inhibition of the local error, it excites the attempts memory trace (\textit{a(t)} term in Eqn. \eqref{eq:mt}) and leads to its update. To resume, no activation occurs within the attempts MT at the beginning of learning, meaning that there is no inhibition toward the LC:phasic NF. After several trials and if the error remains high, inhibition will prevent the LC from pursuing the learning of that particular goal in the immediate future. However, the inhibition of a particular goal decreases after focusing on a different goal. By tuning the build-up time ($\tau_+$) of the transient action node, it is possible to modulate the number of motion attempts. If \textit{$\tau_+$} is large, the robot will significantly increase the number of trials to learn a goal. By setting a high value for \textit{$\tau_-$} (attempts MT), the error level will decay more slowly. This would extend the time scale in which a goal can be selected again. Finally, the force node is activated under excitation of the threshold node and will inhibit the IOR and the activation node in the LC to avoid raising the resting level of the slow boost component within the LC. This last inhibition essentially delays for a short time the next goal selection while the IOR activations within the action module are not completely down. 

Regarding the action module, one of the technical novelties rests in the use of the convolution applied to the object, as well as the inhibition of return. This agency provides a simple and efficient way to generate poses around the object and provides a method to focus the random exploration \cite{wilson2020deep}. As for the persistence mechanism, a key novelty is to provide a mechanism that could adapt depending on the learning behavior. The build-up of activation within the attempts NF is similar to the habituation mechanism and can be seen as a habituation to prediction error, on the difference that the prediction error level continuously changes during the learning of a goal. The use of dynamic neural fields supports the flexibility to modulate the number of attempts made by the robot, as well as the inhibition of the goal for future selection if the error remains significant. As a result, it is possible to observe diverse learning behavior by tuning the parameters of these neural fields (see sec. \ref{res_persist})

\subsection{Curiosity}\label{im}

\subsubsection*{Error module}
When the robot discovers a new goal, a forward and an inverse model are created. The forward model outputs the first error to compute the learning progress with dynamic neural fields (Fig. \ref{fig:general}). More precisely, the error controller forms a two-dimensional Gaussian input located at the location of the goal angle and object, with an amplitude corresponding to the error value (goal error MT in the Error component). During exploitation, the robot performs a motion and observes the outcome in the environment. In the absence of changes (i.e., the object remains stationary), the error is unchanged. In case of any changes, the error controller computes the error with the forward model. Here, the model follows a precise timing for the error computation, where the observation takes place ahead of the end of the motion. In practice, the robot pushes the object, computes the error while going back to a resting position, and finally signals that the motion is done. The curiosity mechanism is then divided into two parts: the error and learning progress processes. The former focuses on characterizing the error of a goal as neural activity, while the latter computes the error's level dynamically to deliver the learning progress.

The compute error NF gathers error levels from all the goals. When the robot pursues the learning of a goal, the error level of the goal error MT is combined with the filter MT to affect only the current goal. Indeed, this last field regroups each error and inhibits the current goal, whereas the goal error MT provides a single peak (i.e., the goal error's activation). It is worth mentioning that the activation function (\textit{$\sigma$}) used for this field, local goal NF, local error NF, and updated error NF is a ReLU function. The time node acts as a trigger when a new error value is computed. Once a new error is projected by the error controller, this brief activation supports the memory trace update as well as boosting the resting level of other neural fields. The errors MT keeps in memory each goal value and only updates them once the end-of-action node (EOA) is active. The last main fields are the local goal NF and the updated error NF. They have a resting level of $-2$, meaning that they yield peaks of activation only when they receive the boost from the time node. As mentioned earlier, the error is computed before the robot reaches its resting pose and activates the EOA node. This implies that when the time node is active, the local error NF contains the error's goal \textbf{before} the update, whereas the updated error NF delivers the newly calculated value. These last fields prepare the future neural computation of the various learning progresses.

\subsubsection*{Learning progress component}
The learning progress mechanism is straightforward (Fig. \ref{fig:general}). It receives inputs from the updated error NF and the local goal NF that are projected onto zero-dimensional nodes with a resting level of $-1.025$. The time node remains the same as the error architecture. With this pattern of incoming activation, the rise and low nodes are responsible for detecting if the updated error value is, respectively, superior or inferior to the previous level of activation (local goal NF). The filter NF serves the same purpose as the one described in the error architecture, providing the error level of each goal except the current one. Then, the compute LP memory trace is where the learning progress of a goal is computed. If the low node is active, the suprathreshold activation from the variation NF will project into compute LP and build up an activation. When the rise node is active, it indicates that the error level is rising. In that case, there will be no activation within the variation NF, and the compute LP neural field will decay the learning progress of the current goal. If an activation was present within the variation NF, the activation will increase depending on the value of (\textit{$\tau_+$}). Finally, there is a case where the error values before and after the update are too close to each other, indicating that the progress is not significant. Then, the rise and low nodes will not activate. The specific resting level value of these nodes supports the fine-tuning of this specific event. Here, there will be no activation if the error does not vary beyond a threshold of 0.025.

\section{Experimental setup}\label{exp}

The purpose of the experiments is to evaluate the learning behavior of the architecture under different parameters and with different objects. More specifically, we want to determine the role of habituation, persistence, and error inhibition during goal discovery and learning. We will also analyze the tonic and phasic activation within the LC. Then, we will assess how the Locus Coeruleus model that gathers bottom-up attention and curiosity leads to an open-ended learning approach with a decentralized cognitive control of goals.

\begin{figure}[ht]
\centering
\subfloat{\includegraphics[width=1.65in]{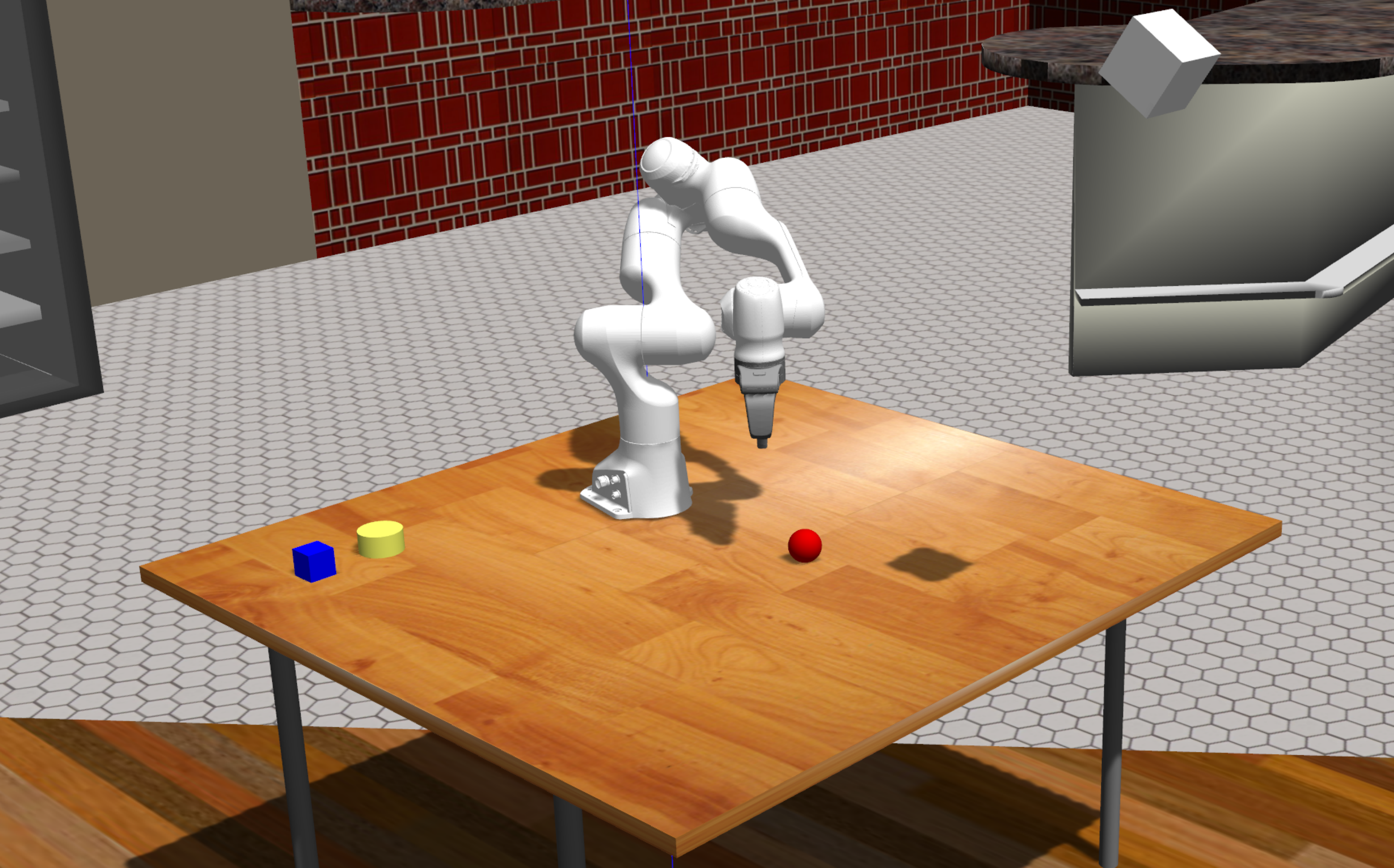}%
\label{fig:scene_robot}}
\hfil
\subfloat{\includegraphics[width=1.805in]{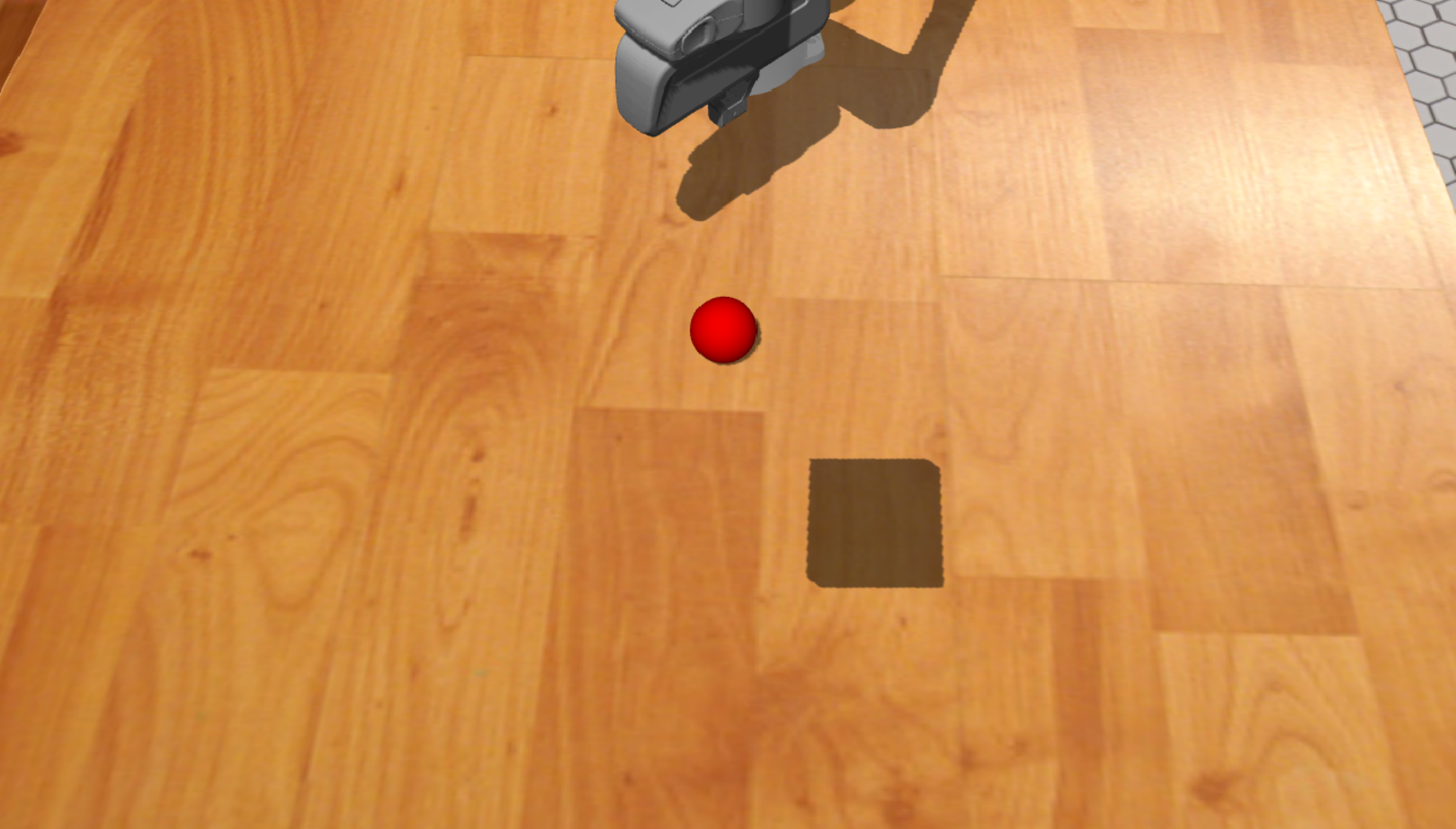}%
\label{fig:camera_robot}}
\caption{(Left) The overall scene of the experiment setup with the Franka Panda in a resting position. The polygon hovering over the table is the virtual camera. (Right) Field of view of the camera with the red ball in sight.\label{fig:scene}}
\end{figure}
The scene simulates a Franka Emika collaborative robot in Gazebo (Fig. \ref{fig:scene}). Dynamic neural fields are simulated with the Cedar software \cite{lomp2013software}, and the ROS middleware \cite{ros} assures communication between the robot and the neural fields. The objects used for the setup are a red ball, a blue cube, and a yellow cylinder. We specifically choose these objects with regard to their difficulty in manipulating them. Since the end-effector orientation is fixed with the gripper closed, it is relatively easy to push the cube in different directions. For the cylinder, the last pose generated has to be more precise to avoid seeing the object slip away during a motion. The ball is the most difficult object to learn to manipulate, since a slight error in the end effector position can push the ball in various directions. The experiment begins when an object is present in the visual field of the camera. The object is respawned in the center of the table if it is outside the reach of the robot. This means that objects can be moved from various locations on the table as long as they are not out of reach. Each neural field is bounded between 0 and 100 along its respective dimension. The gripper orientation is fixed, as well as the dimension z when performing a motion. During the goal discovery stage, the end effector position of the robot is recorded from the resting state until it reaches the goal pose (\ref{arch}). The number of recorded poses remains almost the same, depending on the motion ($\approx$ 40 poses), and they are used to generate a DMP if the motion leads to new stimuli.

\paragraph*{Perception}
Visual perception is processed by a virtual camera with a resolution of 1280$\times$720. This camera is calibrated to the robot space coordinates to transform the image of an object from the camera point of view to the robot frame. A tracker identifies and filters the objects by color and returns the filtered image to the position object NF in the action module. The result is a neural activation that represents the object within the robot's space coordinate system. At the same time, the color tracker returns a continuous input to several neural fields (perception and habituation modules). As mentioned earlier, a two-dimensional goal is formalized within a neural field by collecting the motion direction and color of the object along the horizontal and vertical dimensions, respectively.

\paragraph*{Learning}
The learning of a goal is formalized by a forward and inverse model through multilayer perceptrons. The forward model has one input layer (4D), one hidden layer (6D) and one output layer (1D). The inputs are the current position of the object (x,y) and the motor command (final pose of the end effector $x,y$). Then, the output of this neural network is the motion direction (goal angle) of the object. The inverse model has one layer input (3D) composed of the current object position (x,y) and the goal angle of the object. After the hidden layer (4D), the output (2D) is the motor command (end effector position on x and y). Each DMP is associated with its own forward and inverse model. The purpose of the inverse model associated with a DMP is to learn to generate the correct end-effector final pose for a given goal and object position. More details of the learning parameters can be found in Appendix \ref{app:training}.

\section{Results}\label{res}

\subsection{Habituation}

\begin{figure}[ht]
\centering
\includegraphics[width=3.25in]{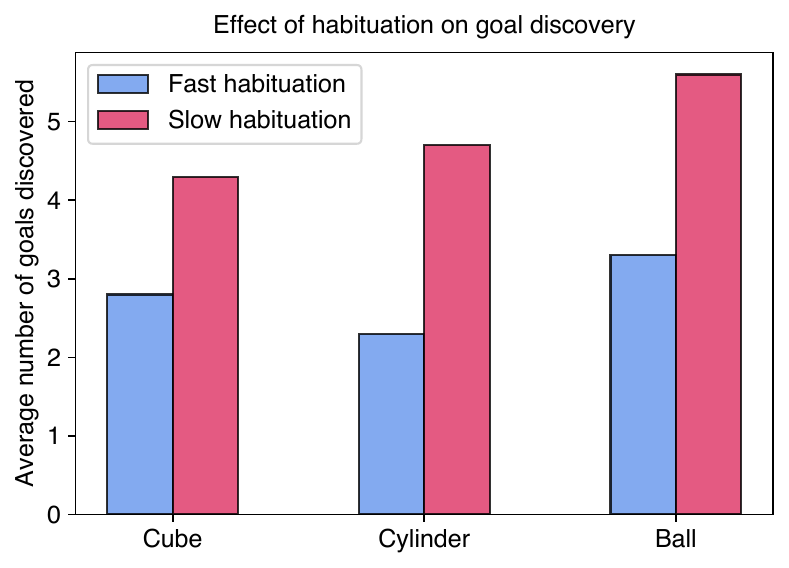}%
\caption{Average number of goals discovered for 10 experiments on each object with fast and slow habituation. The values for fast and slow habituation ($\tau_+$) are 2 and 4 seconds. \label{fig:habit}}
\end{figure}

In this section, we evaluate the effect of habituation during exploration (see Fig. \ref{fig:habit}). To do so, we performed ten explorations for each of the three objects (cube, cylinder, and ball) by varying the \textit{$\tau_+$} parameter of the visual memory trace (Eq. \ref{eq:mt}).

As expected, a slow habituation ($\tau_+ = 4$ seconds) allows the robot to spend more time exploring, leading to an increase in the number of goals discovered (see Fig. \ref{fig:habit}). We can notice a difference among the objects with the ratio :
\begin{equation}
    R_{avg} = S_{avg} / F_{avg}
\end{equation}

\begin{figure*}
%\captionsetup[subfigure]{justification=centering}
\centering
\subfloat{\includegraphics[width=2.3in]{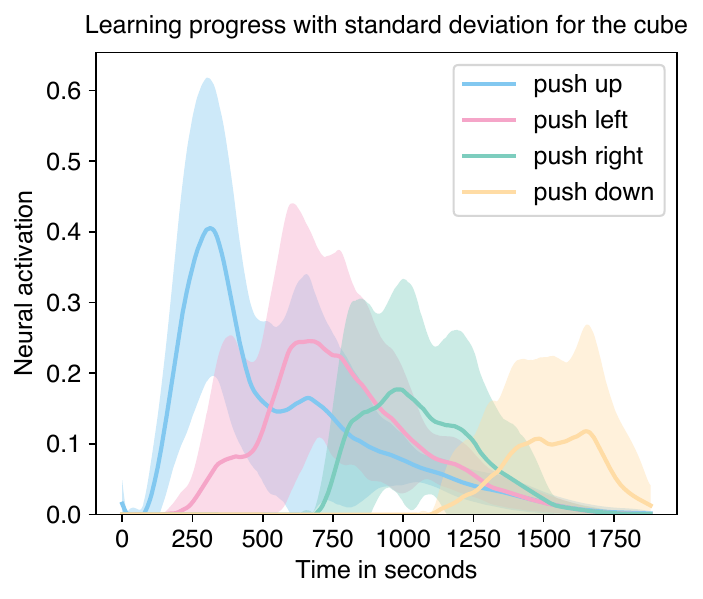}}%
\hfil
%\label{fig:cube_mid}}
\subfloat{\includegraphics[width=2.35in]{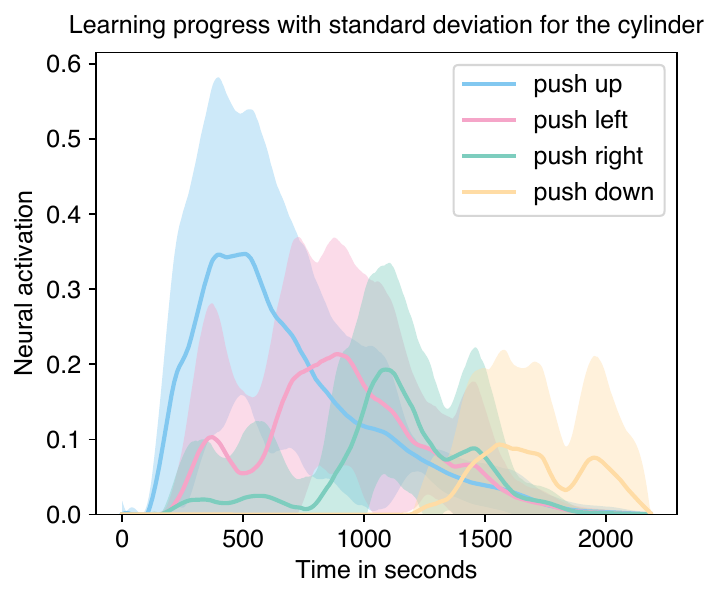}}%
\hfil
%\label{fig:cylinder_mid}}
\subfloat{\includegraphics[width=2.3in]{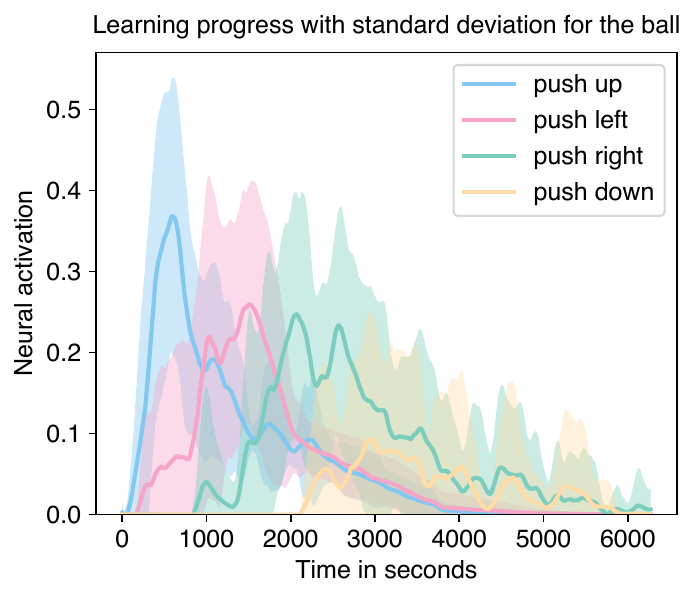}}%
%\label{fig:ball_mid}}
\caption{Neural activation representing the learning progress evolution in time with standard deviation across 20 experiments for the baseline experiment. The figure depicts the average learning progress for the cube (left), the cylinder (middle), and the ball (right).}
\label{fig:progress_baseline}
\end{figure*}
\begin{table*}[tbp]
    \centering
    \begin{tabular}{|*{10}{c|}}
    \hline
    \multicolumn{1}{|c}{} & \multicolumn{3}{|c}{Baseline} & \multicolumn{3}{|c|}{Error Inhibition} & \multicolumn{3}{|c|}{Persistence} \\ \hline
    Goal sequence & Cube & Cylinder &Ball &Cube & Cylinder &Ball &Cube & Cylinder &Ball \\ \hline
    up $\rightarrow$ left & 3.15e-5& 0.03&1.94e-6&\textbf{0.1696}&1.27e-3&1.1e-4&2.83e-5&3.84e-6&1.21e-5 \\ \hline
    left $\rightarrow$ right & 3.85e-9& 1.85e-7&1.94e-6&2.21e-4&1.27e-3&1.33e-3&2.83e-5&3.84e-6&2.12e-4 \\ \hline
    right $\rightarrow$ down & 1.18e-9& 5.73e-7&2.69e-6&2.38e-4&1.99e-3&7.98e-4&3.21e-5&4.3e-6&03.14e-4 \\ \hline
    \end{tabular}
    \caption{Mann-Whitney U test between 2 consecutive goals for each object, under each setting and for all experiments.}
    \label{table:man-whitney}
\end{table*}

where $S_{avg}$ and $F_{avg}$ are the average number of goals discovered for a slow and fast habituation.
The ratios for the cube, cylinder and the ball are, respectively, 1.53, 2.04 and 1.69. This suggests that, given enough habituation time, the robot can double the number of goals discovered with the cylinder. This can be explained by the orientation of the gripper and the inhibition of return. Since the gripper's orientation is fixed, only a limited number of actions can trigger the discovery of a new goal. The movement is guided by the inhibition of return around the object's location, so this inhibition has to be precise and thus requires more trials to generate a meaningful motion. In the following section, we provide an analysis of the learning trajectory of different goals with curiosity.

\subsection{Goal learning progress}\label{}

\subsubsection{Time-course of learning progress between objects}

The evolution of learning progress is presented in Figure \ref{fig:progress_baseline}. The experiment is repeated twenty times for each object, and to compare the learning between objects and parameters, we fixed the same four goals. These goals are : push up, push to the left, push to the right, push down. The initialization of the forward model weights determines the first error during goal discovery, which then influences the future goal selection for learning. We consider this experiment as the baseline, with a persistence value (transient action MT node) at $4500ms$ and $100ms$ for $\tau_+$ and $\tau_-$, respectively. The error inhibition (attempts MT in the persistence module projected on LC:phasic) is settled at $100ms$ for $\tau_+$ and $\tau_-$, which indicates that there is no error inhibition staying in memory between the selection of a goal. The learning progress (LP MT in the learning progress module) is fixed at $2000ms$ for $\tau_-$ and $\tau_+$. Regarding the LP, we arbitrarily choose these values because a high $\tau_-$ would signify that the goal error takes more time to decay even if the error coming from the forward model is already steady at a low level. We performed a Kruskal-Wallis test to evaluate the significance of goal ordering for each object. For the cube, we take the distribution from $t = [0,1800]s$; for the cylinder, we sample $t = [0,2250]s$; for the ball, the distribution range is $t = [0,6000]s$. The goal ordering analysis aims to demonstrate the influence of error inhibition on learning flexibility. We hypothesize that this specific inhibition (i.e., inspired by \cite{ahmadlou_subcortical_2025}) can help disengage from a goal that is too difficult to learn and thus reduce persistence.  

\begin{figure*}
\centering
\subfloat{\includegraphics[width=2.3in]{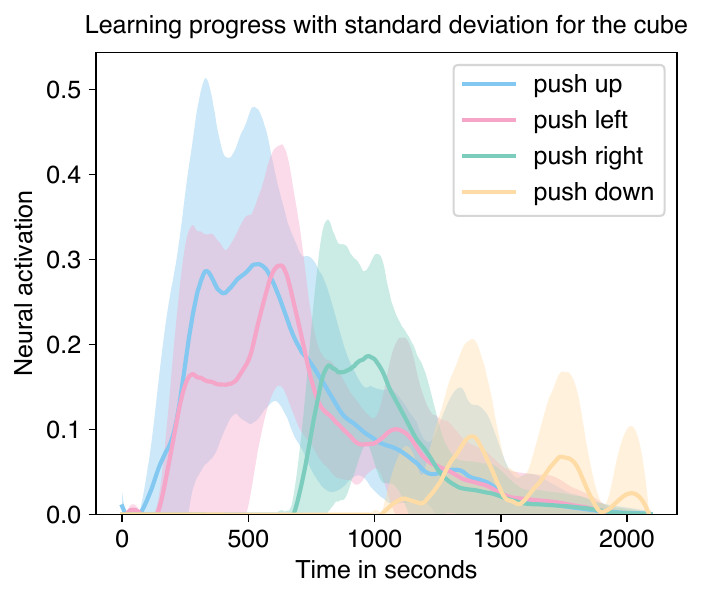}}
\hfil
\subfloat{\includegraphics[width=2.36in]{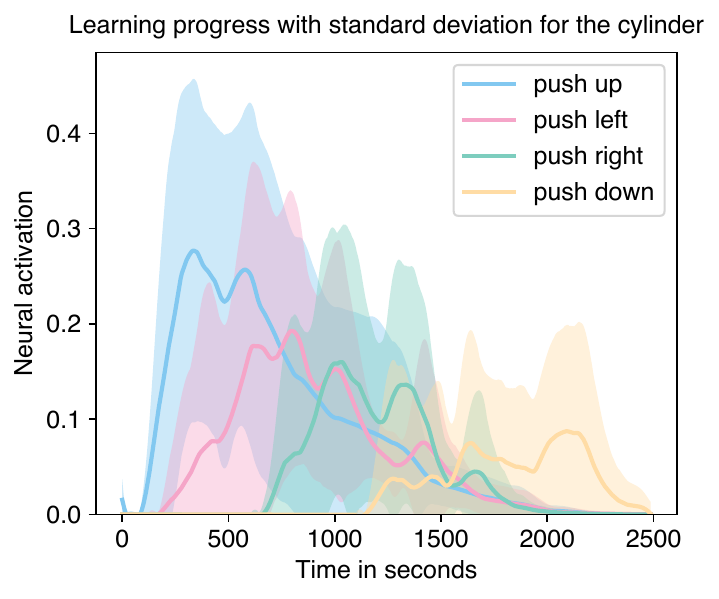}}
\hfil
\subfloat{\includegraphics[width=2.3in]{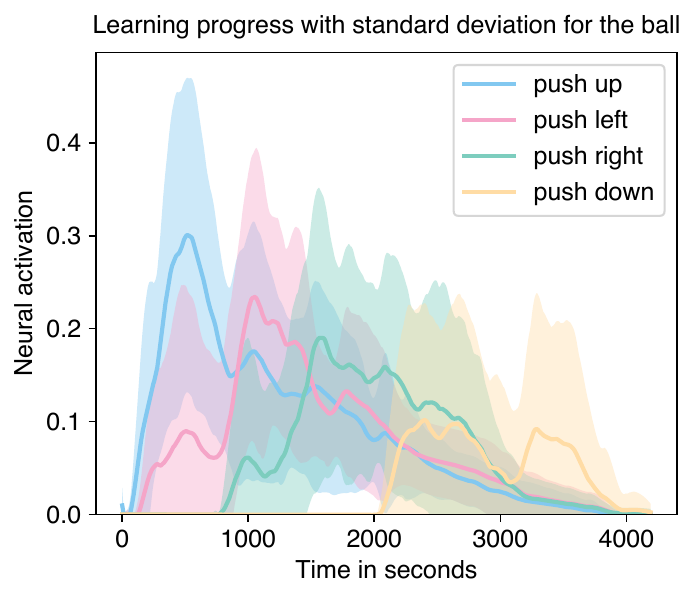}}
\hfil
\subfloat{\includegraphics[width=2.3in]{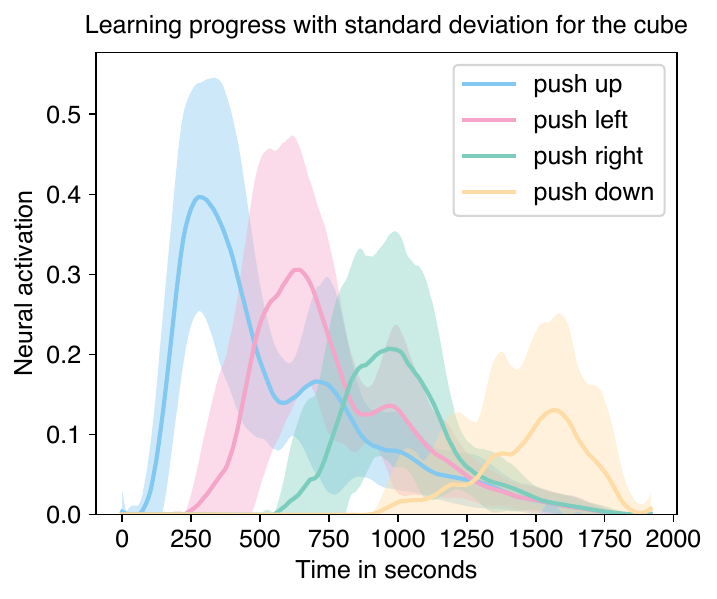}}
\hfil
\subfloat{\includegraphics[width=2.3in]{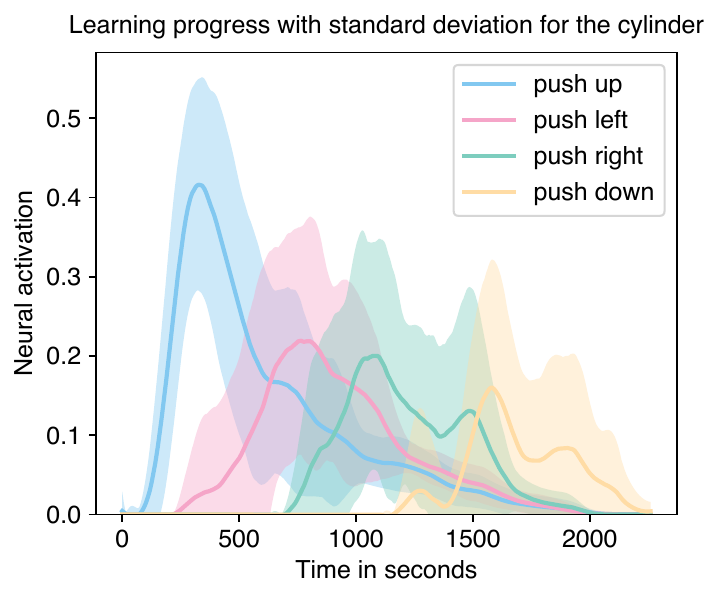}}
\hfil
\subfloat{\includegraphics[width=2.25in]{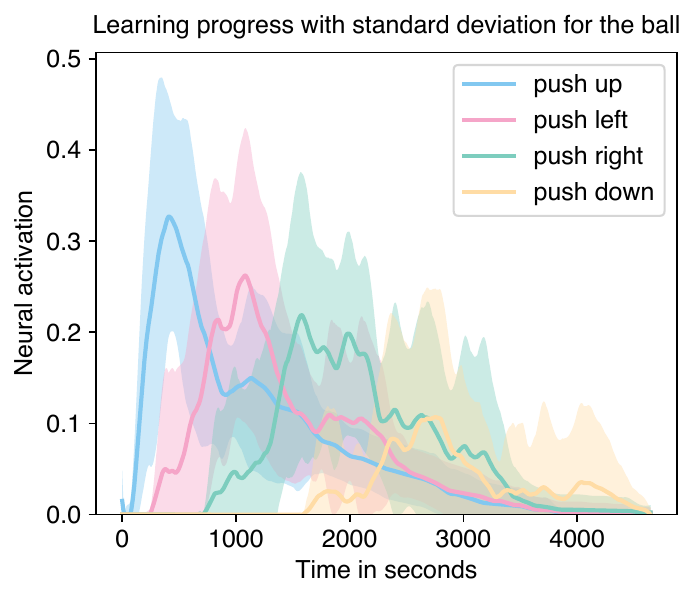}}%
\caption{Top row : goals learning progress across 20 experiments under the error inhibition parameters for the cube (left), the cylinder (middle) and the ball (right). Second row : goals learning progress across 20 experiments under the persistence parameters for the cube (left), the cylinder (middle) and the ball (right).}
\label{fig:progress_inh}
\end{figure*}

The distribution differences between the four goals are significant with $H(4)$, $p < 0.01$ among the three objects. Even if the sequence of the four goals is significant among all the runs, we still perform a Mann-Whitney U test (see Tab. \ref{table:man-whitney}) between two successive goals (\textit{push-up $\rightarrow$ push-left, push-left $\rightarrow$ push-right, push-right $\rightarrow$ push-down}) to verify a possible overlap between them in certain runs. The statistical test for the cylinder is almost not significant, indicating a different ordering between $up$ and $left$ among several runs. We can conclude that the goals follow the same learning order for the baseline experiment. However, these goals do not reflect a developmental trajectory, as they can be learned independently. Moreover, learning begins with the goal that has the highest error, which depends on the initialization of the synaptic weights of the forward model. The difficulty of a goal to be learned can be evaluated by the time spent until the activation of the learning progress falls below $0.01$. For the ball, it is considerably more difficult to learn the set of goals than for the cylinder and the cube ($\approx$ 6000s). This is essentially due to the precision required to generate a $goal$ (Appendix \ref{app:dmp}).

\subsubsection{Winner inhibits similar goals}

\begin{figure}[ht]
\centering
\subfloat{ 
%\hspace{2mm}
\includegraphics[width=1.65in]{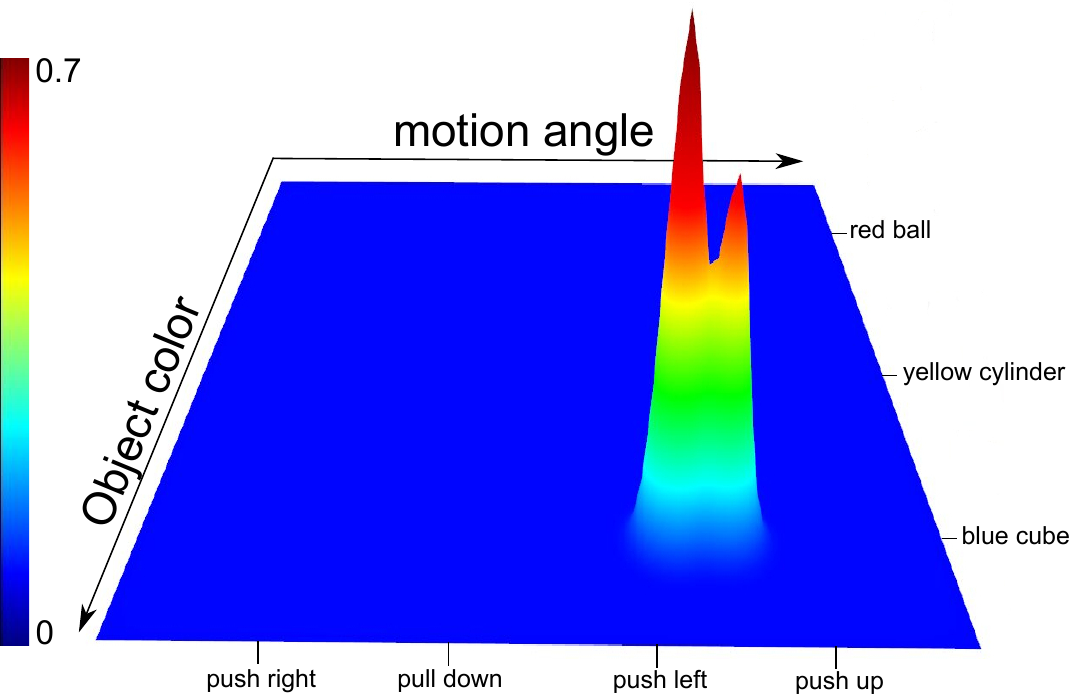}}
\hfil
\subfloat{
%\hspace{0mm}6
\includegraphics[width=1.65in]{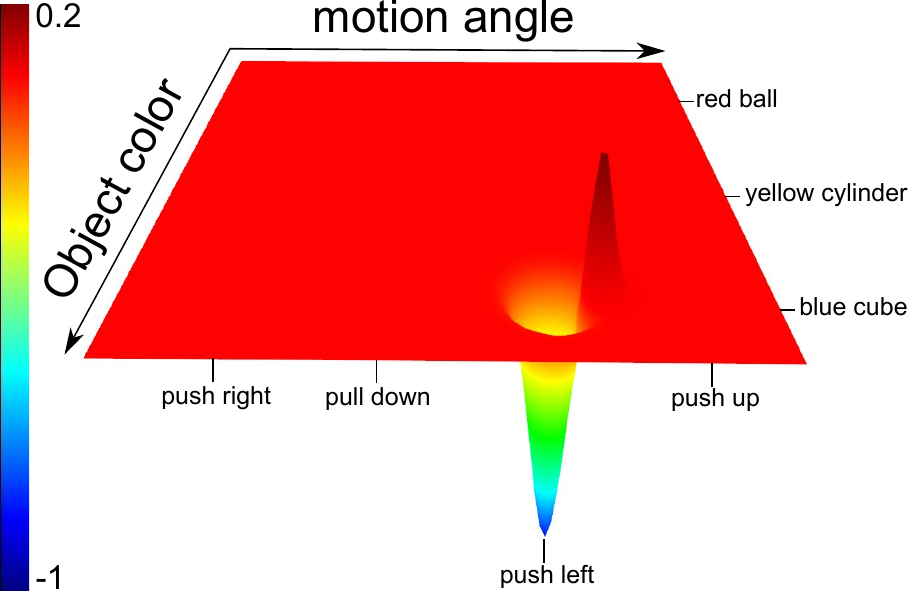}}
\caption{(Left) Two similar goals in errors MT (Error mechanism) after goal discovery. (Right) Same two goals in filter MT, but the selected goal for learning is inhibited to compute the error. \label{fig:similar_goals}}
\end{figure}

During exploration with the habituation stage, it is possible for the robot to discover similar goals (see Fig. \ref{fig:similar_goals}). 

For example, the robot can push the cube to the left and push the cube to the left, but slightly shifted by a few degrees upward. In that case, only the goal with the highest error from the forward model will be learned in the future. This comes from the neural dynamics of the filter MT (error module) and, more particularly, from the inhibition of the current goal (see Fig. \ref{fig:similar_goals}). In fact, the filter MT represents the various discovered goals except the current goal via inhibition of current goal NF. This localized inhibition not only affects the goal peak, but also partially or completely alters any peaks of activation in proximity. This effect depends on the standard deviation ($\sigma$) of the inhibition. A nearby goal will be affected by having its neural activation lowered. This results in a lower neural activation for the similar goal when the current goal sees its error updated. An elevated standard deviation for each neural activation within the neural fields in the Error module would mean that the goals must significantly differ from each other to be learned. The size and amplitude of suprathreshold activations have been shown to have an impact on the learning of sensorimotor contingencies \cite{houbreinh}. This effect can be seen as a way to focus first on acquiring a diverse set of goals before learning only slightly different goals later on. The consequences of this effect lead to developmental questions and should be investigated more deeply.

\subsection{Influence of inhibition and persistence}\label{res_persist}

After learning goals under the baseline parameters, we investigate the impact of error inhibition through the attempts MT in the persistence mechanism. We settle the time parameters at $\tau_+ = 2000ms$ and $\tau_- = 1500ms$ and run the experiment twenty times for each object (see Fig. \ref{fig:progress_inh}, top row). We selected these parameters to ensure that the inhibition develops and dissipates more slowly. When a goal error proves challenging to reduce, it induces mild inhibition, preventing the goal from being chosen again in the near future. We repeated a Kruskal-Wallis test for the cube, cylinder, and ball for the time intervals $t = [0,2100]s$, $t = [0,2500]s$ and $t = [0,4250]s$, respectively. The distribution of the four goals in the 20 experiments was conclusive with $H(4), p < 0.01$. A Mann-Whitney test was performed to determine whether the goals consistently follow the same sequence (see Table \ref{table:man-whitney}). We observed that the test is not significant for the cube, which means that there is no clear ordering between $pushing-up$ and $pushing-left$. If the learning progress for the goal $pushing-up$ remains stagnant after a learning session, inhibition will hinder near-future selection of this goal. The statistical test result was significant for the cylinder and the ball, but with a higher value than the baseline experiment. This demonstrates the presence of several runs for both objects, where the goals do not follow the mainstream sequence pattern. If we compare the inhibition settings with the baseline, we can see an increase in the learning time for the cube and the cylinder. However, this learning time is shorter for the ball, and the goals $push-right$ and $push-down$ are less subject to fluctuation. For the ball baseline experiment, $push-right$ and $push-down$ take, respectively, $5000s$ and $3900s$ against $3000s$ and $2000s$ under the inhibition parameters. The baseline experiment relies only on the learning progress of the goals to continue the exploitation, which provokes only minimal learning fluctuation in the case of easy objects to learn (cube and cylinder). For a difficult object, the system has to continue learning challenging goals exhibiting a certain learning progress with no possibility to select a different goal. Under the inhibition parameters, the model introduces more flexibility in goal selection since learning progress is not the only determining factor. Essentially, this suggests that the robot does not finish learning a goal entirely before learning a new one. This switch between goals indicates an advantage on the ball, whereas it introduces more fluctuation for the cube. Apart from a certain extension of learning time ($\approx 250s$), the cylinder does not appear to be affected by the inhibition parameters.

We introduce the persistence settings by modifying the transient action MT node ($\tau_+ = 6000ms$, $\tau_- = 100ms$). These settings extend the learning time before the LC module disengages from learning the current goal. This parameter is analogous to the exploration time seen in studies implementing intrinsically motivated systems \cite{seepanomwan_intrinsic_2020}, \cite{sener_exploration_2021}. Each object was subjected to the experiment twenty times, followed by a Kruskal-Wallis test to evaluate the goal distributions. The test was conclusive for the three objects with $H(4), p < 0.01$, and the Mann-Whitney U test between two consecutive goals does not show a significant variation in the learning sequence for all of them (see Table \ref{table:man-whitney}). However, statistical significance is closer to the baseline experiment than for the inhibition parameters. The persistence parameters provide results similar to the baseline for the cube. With respect to the cylinder, the distributions of the different learning progresses are more distinct. This manifests itself as less learning overlap between the goals in all the runs. With regard to the ball, the learning time for all goals is reduced compared to baseline but increased compared to inhibition.

Investigating three distinct objects of varying learning complexity enables assessing error inhibition and persistence. Error inhibition triggers disengagement from a goal too difficult to learn (i.e., no evolution of the learning progress) and reorients learning to a different goal. Persistence determines the number of trials to learn a goal before disengagement. Together, these parameters modulate the learning of the goals in a flexible way based on the complexity of the objects. Here, the baseline experiment is optimal for the cube with no significant learning fluctuation. The ball exhibits the shortest learning time with moderate learning fluctuation under the inhibition parameters, and the cylinder displays a weaker overlap of learning between goals under the persistence criterion.

%\begin{figure*}
%\centering
%\subfloat{ 
%\includegraphics[width=4in]{time_course.png}}
%\hfil
%\subfloat{
%\includegraphics[width=3in]{tonic_phasic_final.png}}
%\caption{(Left) time course of object discovery and learning by introducing the cube first, the cylinder in second, and the ball last. Forward model error and learning progress depict the learning. The green curve represent the activation of the explore node. (Right) Time course of the tonic and phasic activation for the discovery and learning of a single goal.}
%\label{fig:attention}
%\end{figure*} 

\subsection{Tonic and phasic activation}
\begin{figure}[h]
\centering
\hspace{2mm}
\includegraphics[width=3.25in]{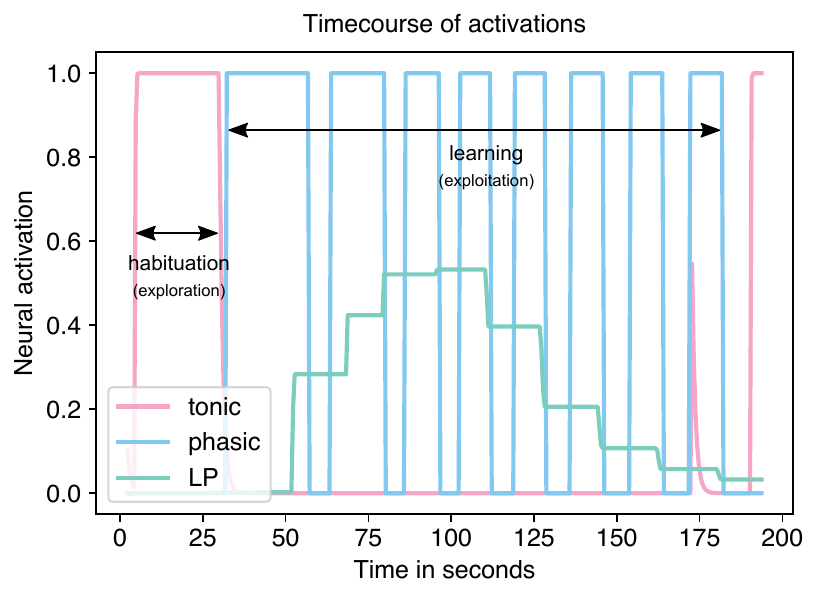}%
\caption{Time course of the tonic and phasic activation for the discovery and learning of a single goal.}
\label{fig:tonic}
\end{figure}
The slow boost component is the core of tonic and phasic activation within the Locus Coeruleus model. In Figure \ref{fig:tonic}, we monitored activations within the tonic and phasic NF. During exploration, the tonic activation remains sustained as long as the robot is not habituated to the object. After learning begins, the slow boost component is periodically reset and provides excitation to the tonic and phasic NF. The boost reset rate depends on the persistence mechanism. Our model proposes an indirect contribution of the error through the persistence mechanism. Indeed, the persistence architecture modulates these periods depending on the goal error (see Appendix \ref{app:persistence}, Figures \ref{fig:sim_persist} and \ref{fig:sim_long_persist} - bottom row). When the error of a goal is substantial, the interval between successive selections increases. In the first goal selection for the baseline experiment, the boost delivers excitation for 16 seconds before dropping down. At the end of learning the same goal, the boost provides excitation for 6.5 seconds. The transient action MT node also influences the period of boost activation. The parameter $\tau_+$ and the frequency of selecting a goal are inversely related. If $\tau_+$ is large, then the selection of goals occurs less often. Moreover, error inhibition leads to more frequent disengagement if there is no learning progress. In our model, the phasic activation of LC results from the goal selection process and the frequency with which the robot switches the learning between different goals.

\subsection{A loop for incremental learning}

\begin{figure}[h]
\centering
\hspace{2mm}
\includegraphics[width=3.4in]{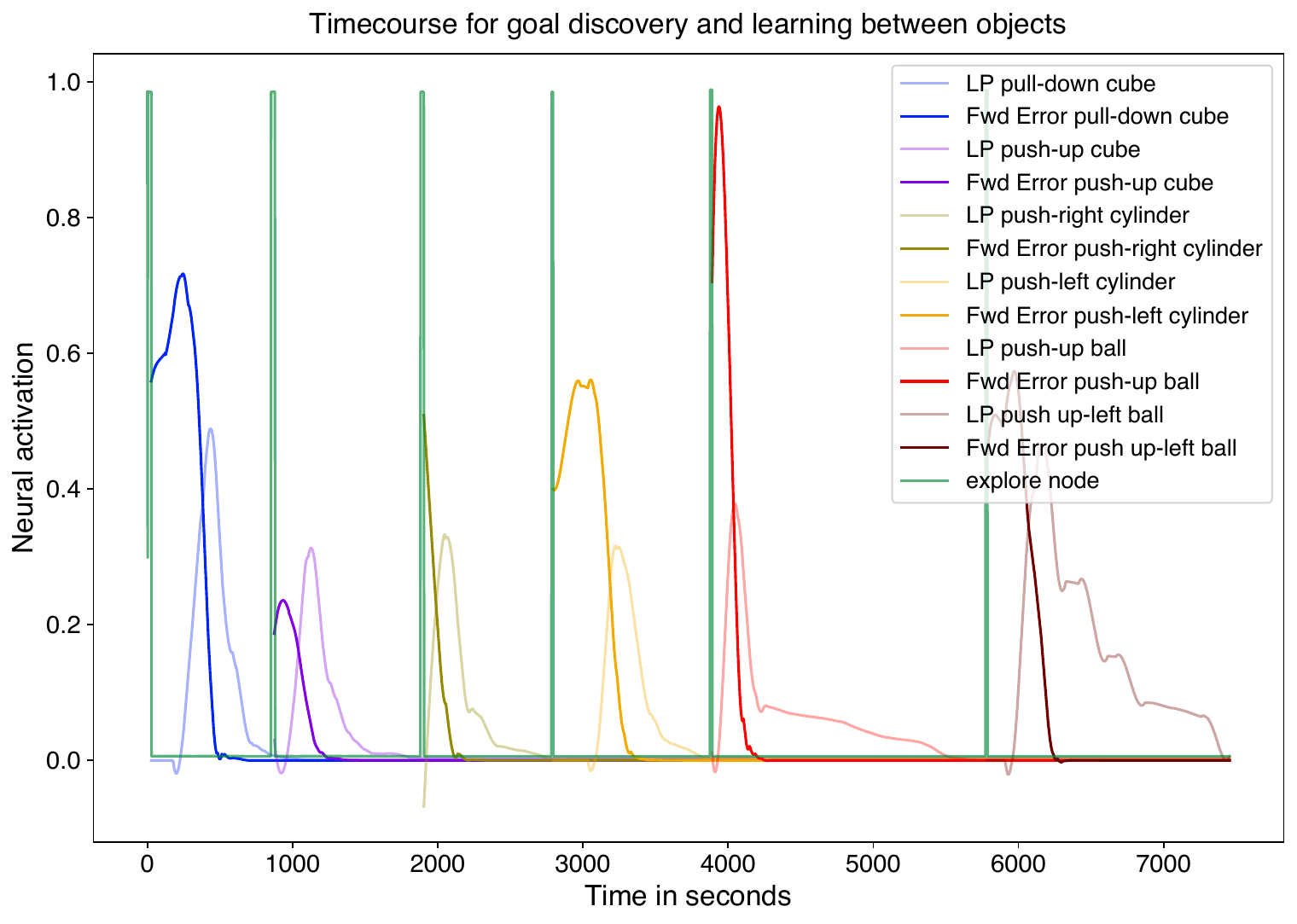}%
\caption{Timecourse of object discovery and learning by introducing the cube first, the cylinder in second, and the ball last. Forward model error and learning progress depict the learning trajectory.}
\label{fig:attention}
\end{figure}

The architecture detects bottom-up events, such as the presence of new objects in the scene, and begins to explore through goal discovery. Here, we reduce the habituation time ($\tau_+$ in visual memory trace), so that each object has a short exploration. We begin by introducing the cube, let the robot discover and learn a goal, then reproduce this step for the cylinder and the ball. This protocol exhibits the properties of the attentional system to focus on new stimuli. The time course of the experiment with the learning progress and forward model error is detailed in Figure \ref{fig:attention}. It is also possible to drive the robot into exploration again after habituation to a specific location by increasing the object stimuli. In practice, it is sufficient to introduce a supplementary excitation within the object selection NF in the habituation mechanism.

\section{Conclusion and Discussion}\label{disc}

The autonomous discovery and learning of goals in robotics is a complex issue to address, and the state-of-the-art represents a wide spectrum of approaches. Here, we take inspiration from neuroscience by rooting curiosity with attention in order to allow a robotic arm to discover and learn different goals. More precisely, we propose a simple model of the Locus Coeruleus, a nucleus in the brain that controls to some extent the switch between exploration and task engagement through the influence of curiosity, habituation, and persistence. At first, the tonic activation within the LC drives the robot to generate actions that will produce new stimuli, and thus discover new goals via bottom-up attention. Then, the curiosity mechanism directly influences the Locus Coeruleus activation with a phasic mode by engaging in learning, depending on both forward model error and learning progress of these goals. In addition, we introduce new cognitive mechanisms to regulate the discovery and learning of goals with, respectively, a habituation and persistence component. The generation of robot motion is performed by inhibition of return and dynamic movement primitives. The complete architecture is modeled with dynamic neural fields (DNFs) except for the forward and inverse models, which are multilayer perceptrons. The use of DNFs supports several technical novelties, such as the generation of poses around an object and the interaction between goals in memory. In addition, the goal error modeling allows widespread connections that bootstrap the shift between goal discovery and learning, as well as control the learning of goals. Regarding the habituation paradigm driving exploration, we demonstrated that the robot was able to discover various goals, with a higher average number of goals for slow habituation. For persistence, we tuned the time spent learning a goal as well as the strength of the error inhibition to the LC and demonstrated a clear influence on learning. Varying these parameters manifests positive and negative impacts on learning that suggest individual optimization depending on the objects. Moreover, the curiosity module exhibits an interesting property in the case of two similar goals, where the learning of one directly inhibits the other. Finally, the results indicate that the architecture continuously oscillates between goal discovery through bottom-up attention and goal learning via curiosity.

However, the current model suffers from several limitations. The objects in the habituation mechanism are distinguished by color, which is not realistic. In the future, we plan to use a variational autoencoder (VAE) to integrate more features and improve the discovery of goals, as demonstrated by researchers \cite{houbreexpl}. To do so, an object and a goal will be characterized by more features (3D object pose, touch sensor, rotation), so a more diverse set of goals can be learned. Regarding exploration, we will use the latent space as a supplementary method to discover new goals through direct exploration \cite{wilson2021balancing}. This way, we will investigate the balance between direct exploration (VAE) and random exploration with the inhibition-of-return method. The orientation of the end effector and the $z$ dimension of the robot were fixed, and we intend to eliminate this limitation to perform a diverse range of robot actions that could lead to different outcomes and potentially produce a developmental trajectory.

Finally, we will explore the tuning of the persistence and error-inhibition parameters together. In fact, the model offers several possibilities, and different learning patterns could still be observed. We would like to investigate whether the number of goals discovered during habituation can be used to predict the parameters for optimal learning, and how to autonomously determine and set these parameters. Finally, the goals discovered were on the same developmental level, and the first learning selection entirely depended on the weights initialization of the forward models. We intend to investigate the possibility of autonomously tuning the persistence and error-inhibition parameters to produce optimal learning based on the object difficulty.

\bibliographystyle{IEEEtran}
\bibliography{literature.bib}

% Generated by IEEEtran.bst, version: 1.14 (2015/08/26)
\begin{thebibliography}{10}
\providecommand{\url}[1]{#1}
\csname url@samestyle\endcsname
\providecommand{\newblock}{\relax}
\providecommand{\bibinfo}[2]{#2}
\providecommand{\BIBentrySTDinterwordspacing}{\spaceskip=0pt\relax}
\providecommand{\BIBentryALTinterwordstretchfactor}{4}
\providecommand{\BIBentryALTinterwordspacing}{\spaceskip=\fontdimen2\font plus
\BIBentryALTinterwordstretchfactor\fontdimen3\font minus
  \fontdimen4\font\relax}
\providecommand{\BIBforeignlanguage}[2]{{%
\expandafter\ifx\csname l@#1\endcsname\relax
\typeout{** WARNING: IEEEtran.bst: No hyphenation pattern has been}%
\typeout{** loaded for the language `#1'. Using the pattern for}%
\typeout{** the default language instead.}%
\else
\language=\csname l@#1\endcsname
\fi
#2}}
\providecommand{\BIBdecl}{\relax}
\BIBdecl

\bibitem{oudeyer_intrinsic_2007}
P.~Y. Oudeyer, F.~Kaplan, and V.~V. Hafner, ``Intrinsic motivation systems for
  autonomous mental development,'' \emph{IEEE Transactions on Evolutionary
  Computation}, vol.~11, no.~2, pp. 265--286, 2007.

\bibitem{oudeyer2010intrinsically}
P.-Y. Oudeyer, A.~Baranes, and F.~Kaplan, ``Intrinsically motivated exploration
  for developmental and active sensorimotor learning,'' in \emph{From motor
  learning to interaction learning in robots}.\hskip 1em plus 0.5em minus
  0.4em\relax Springer, 2010, pp. 107--146.

\bibitem{oudeyer2013intrinsically}
------, ``Intrinsically motivated learning of real-world sensorimotor skills
  with developmental constraints,'' in \emph{Intrinsically motivated learning
  in natural and artificial systems}.\hskip 1em plus 0.5em minus 0.4em\relax
  Springer, 2013, pp. 303--365.

\bibitem{baranes2010}
A.~Baranes and P.-Y. Oudeyer, ``Intrinsically motivated goal exploration for
  active motor learning in robots: A case study,'' in \emph{2010 IEEE/RSJ
  International Conference on Intelligent Robots and Systems}, 2010, pp.
  1766--1773.

\bibitem{BARANES2013}
------, ``Active learning of inverse models with intrinsically motivated goal
  exploration in robots,'' \emph{Robotics and Autonomous Systems}, vol.~61,
  no.~1, pp. 49--73, 2013.

\bibitem{richards1985development}
J.~E. Richards, ``The development of sustained visual attention in infants from
  14 to 26 weeks of age,'' \emph{Psychophysiology}, vol.~22, no.~4, pp.
  409--416, 1985.

\bibitem{atkinson1992changes}
J.~Atkinson, B.~Hood \emph{et~al.}, ``Changes in infants' ability to switch
  visual attention in the first three months of life,'' \emph{Perception},
  vol.~21, no.~5, pp. 643--653, 1992.

\bibitem{posner_inhibition_1985}
M.~Posner, R.~Rafal \emph{et~al.}, ``Inhibition of return: {Neural} {Basis} and
  {Function},'' \emph{Cognitive Neuropsychology}, vol. Vol. 2, pp. 211--228,
  Sep. 1985.

\bibitem{posner1971components}
M.~I. Posner and S.~J. Boies, ``Components of attention.'' \emph{Psychological
  review}, vol.~78, no.~5, 1971.

\bibitem{maclean2009interactions}
K.~A. MacLean, S.~R. Aichele \emph{et~al.}, ``Interactions between endogenous
  and exogenous attention during vigilance,'' \emph{Attention, Perception, \&
  Psychophysics}, vol.~71, no.~5, pp. 1042--1058, 2009.

\bibitem{chica_effects_2009}
A.~B. Chica and J.~Lupiáñez, ``\BIBforeignlanguage{en}{Effects of endogenous
  and exogenous attention on visual processing: {An} {Inhibition} of {Return}
  study},'' \emph{\BIBforeignlanguage{en}{Brain Research}}, vol. 1278, pp.
  75--85, Jun. 2009.

\bibitem{aston-jones_integrative_2005}
G.~Aston-Jones and J.~D. Cohen, ``\BIBforeignlanguage{eng}{An integrative
  theory of locus coeruleus-norepinephrine function: adaptive gain and optimal
  performance},'' \emph{\BIBforeignlanguage{eng}{Annual Review of
  Neuroscience}}, vol.~28, pp. 403--450, 2005.

\bibitem{gottlieb2016motivated}
J.~Gottlieb, M.~Lopes, and P.-Y. Oudeyer, ``Motivated cognition: Neural and
  computational mechanisms of curiosity, attention, and intrinsic motivation,''
  in \emph{Recent developments in neuroscience research on human
  motivation}.\hskip 1em plus 0.5em minus 0.4em\relax Emerald Group Publishing
  Limited, 2016.

\bibitem{ranjbar2020dopamine}
Y.~Ranjbar-Slamloo and Z.~Fazlali, ``Dopamine and noradrenaline in the brain;
  overlapping or dissociate functions?'' \emph{Frontiers in molecular
  neuroscience}, vol.~12, p. 334, 2020.

\bibitem{sophian_habituation_1980}
C.~Sophian, ``Habituation is not enough : novelty preferences, search, and
  memory in infancy,'' \emph{Merrill-Palmer Quarterly of Behavior and
  Development}, vol.~26, no.~3, pp. 239--257, 1980.

\bibitem{schoner2016dynamic}
G.~Sch{\"o}ner and J.~P. Spencer, \emph{Dynamic thinking: A primer on dynamic
  field theory}.\hskip 1em plus 0.5em minus 0.4em\relax Oxford University
  Press, 2016.

\bibitem{baldassarre_what_2011}
G.~Baldassarre, ``What are intrinsic motivations? {A} biological perspective,''
  in \emph{2011 {IEEE} {International} {Conference} on {Development} and
  {Learning}, {ICDL} 2011}, vol.~2, 2011, pp. 1--8.

\bibitem{oudeyertypo}
P.-Y. Oudeyer and F.~Kaplan, ``What is intrinsic motivation? a typology of
  computational approaches,'' \emph{Frontiers in Neurorobotics}, vol.~1, 2009.

\bibitem{mirolli_functions_2013}
M.~Mirolli and G.~Baldassarre, ``Functions and mechanisms of intrinsic
  motivations the knowledge versus competence distinction,'' in
  \emph{Intrinsically {Motivated} {Learning} in {Natural} and {Artificial}
  {Systems}}, G.~Baldassarre and M.~Mirolli, Eds.\hskip 1em plus 0.5em minus
  0.4em\relax Springer Berlin Heidelberg, 2013, vol. 9783642323, pp. 49--72.

\bibitem{oudeyer2016intrinsic}
P.-Y. Oudeyer, J.~Gottlieb, and M.~Lopes, ``Intrinsic motivation, curiosity,
  and learning: Theory and applications in educational technologies,''
  \emph{Progress in brain research}, vol. 229, pp. 257--284, 2016.

\bibitem{schmidhuber1991curious}
J.~Schmidhuber, ``Curious model-building control systems,'' in \emph{Proc.
  international joint conference on neural networks}, 1991, pp. 1458--1463.

\bibitem{lefort_active_2015}
M.~Lefort and A.~Gepperth, ``Active learning of local predictable
  representations with artificial curiosity,'' in \emph{5th {Joint}
  {International} {Conference} on {Development} and {Learning} and {Epigenetic}
  {Robotics}, {ICDL}-{EpiRob} 2015}.\hskip 1em plus 0.5em minus 0.4em\relax
  IEEE, 2015, pp. 228--233.

\bibitem{santucci_grail_2016}
V.~G. Santucci, G.~Baldassarre, and M.~Mirolli, ``{GRAIL}: {A}
  {Goal}-{Discovering} {Robotic} {Architecture} for {Intrinsically}-{Motivated}
  {Learning},'' \emph{IEEE Transactions on Cognitive and Developmental
  Systems}, vol.~8, no.~3, pp. 214--231, Sep. 2016.

\bibitem{seepanomwan_intrinsic_2020}
K.~Seepanomwan, D.~Caligiore \emph{et~al.}, ``Intrinsic {Motivations} and
  {Planning} to {Explain} {Tool}-{Use} {Development}: {A} {Study} with a
  {Simulated} {Robot} {Model},'' \emph{IEEE Transactions on Cognitive and
  Developmental Systems}, vol.~PP, pp. 1--1, May 2020.

\bibitem{laversanne-finot_intrinsically_2021}
A.~Laversanne-Finot, A.~Péré, and P.-Y. Oudeyer, ``Intrinsically {Motivated}
  {Exploration} of {Learned} {Goal} {Spaces},'' \emph{Frontiers in
  Neurorobotics}, vol.~14, p. 109, 2021.

\bibitem{sener_exploration_2021}
M.~I. Sener, Y.~Nagai \emph{et~al.}, ``Exploration with {Intrinsic}
  {Motivation} using {Object}-{Action}-{Outcome} {Latent} {Space},'' \emph{IEEE
  Transactions on Cognitive and Developmental Systems}, pp. 1--1, 2021.

\bibitem{oudeyer_how_2016}
P.-Y. Oudeyer and L.~B. Smith, ``\BIBforeignlanguage{eng}{How {Evolution} {May}
  {Work} {Through} {Curiosity}-{Driven} {Developmental} {Process}},''
  \emph{\BIBforeignlanguage{eng}{Topics in Cognitive Science}}, vol.~8, no.~2,
  pp. 492--502, Apr. 2016.

\bibitem{gottlieb_towards_2018}
J.~Gottlieb and P.-Y. Oudeyer, ``\BIBforeignlanguage{en}{Towards a neuroscience
  of active sampling and curiosity},'' \emph{\BIBforeignlanguage{en}{Nature
  Reviews Neuroscience}}, vol.~19, no.~12, pp. 758--770, Dec. 2018.

\bibitem{verbruggen}
F.~Verbruggen, I.~P.~L. McLaren, and C.~D. Chambers, ``Banishing the control
  homunculi in studies of action control and behavior change,''
  \emph{Perspectives on Psychological Science}, vol.~9, no.~5, pp. 497--524,
  2014.

\bibitem{logan_executive_1985}
G.~D. Logan, ``Executive control of thought and action,'' \emph{Acta
  Psychologica}, vol.~60, no.~2, pp. 193--210, Dec. 1985.

\bibitem{monsell_control_2000}
S.~Monsell and J.~Driver, \emph{Control of {Cognitive} {Processes}: {Attention}
  and {Performance} {XVIII}}.\hskip 1em plus 0.5em minus 0.4em\relax The MIT
  Press, 2000.

\bibitem{di_domenico_emerging_2017}
S.~I. Di~Domenico and R.~M. Ryan, ``The {Emerging} {Neuroscience} of
  {Intrinsic} {Motivation}: {A} {New} {Frontier} in {Self}-{Determination}
  {Research},'' \emph{Frontiers in Human Neuroscience}, vol.~11, p. 145, 2017.

\bibitem{engelmann_motivation_2014}
J.~B. Engelmann and L.~Pessoa, ``Motivation sharpens exogenous spatial
  attention,'' \emph{Motivation Science}, vol.~1, no.~S, pp. 64--72, 2014.

\bibitem{wulf_simply_2003}
G.~Wulf and N.~McNevin, ``Simply distracting learners is not enough: {More}
  evidence for the learning benefits of an external focus of attention,''
  \emph{European Journal of Sport Science}, vol.~3, no.~5, pp. 1--13, Dec.
  2003.

\bibitem{wulf_directing_2001}
G.~Wulf and W.~Prinz, ``\BIBforeignlanguage{en}{Directing attention to movement
  effects enhances learning: {A} review},''
  \emph{\BIBforeignlanguage{en}{Psychonomic Bulletin \& Review}}, vol.~8,
  no.~4, pp. 648--660, Dec. 2001.

\bibitem{poolton_benefits_2006}
J.~M. Poolton, J.~P. Maxwell \emph{et~al.}, ``Benefits of an external focus of
  attention: {Common} coding or conscious processing?'' \emph{Journal of Sports
  Sciences}, vol.~24, no.~1, pp. 89--99, Jan. 2006.

\bibitem{peelen_endogenous_2004}
M.~V. Peelen, D.~J. Heslenfeld, and J.~Theeuwes,
  ``\BIBforeignlanguage{en}{Endogenous and exogenous attention shifts are
  mediated by the same large-scale neural network},''
  \emph{\BIBforeignlanguage{en}{NeuroImage}}, vol.~22, no.~2, pp. 822--830,
  Jun. 2004.

\bibitem{chica_two_2013}
A.~B. Chica, P.~Bartolomeo, and J.~Lupiáñez, ``\BIBforeignlanguage{en}{Two
  cognitive and neural systems for endogenous and exogenous spatial
  attention},'' \emph{\BIBforeignlanguage{en}{Behavioural Brain Research}},
  vol. 237, pp. 107--123, Jan. 2013.

\bibitem{di2014role}
D.~Di~Nocera, A.~Finzi \emph{et~al.}, ``The role of intrinsic motivations in
  attention allocation and shifting,'' \emph{Frontiers in psychology}, vol.~5,
  p. 273, 2014.

\bibitem{baldassarre2019embodied}
G.~Baldassarre, W.~Lord \emph{et~al.}, ``An embodied agent learning affordances
  with intrinsic motivations and solving extrinsic tasks with attention and
  one-step planning,'' \emph{Frontiers in neurorobotics}, vol.~13, p.~45, 2019.

\bibitem{wilson2020deep}
R.~Wilson, S.~Wang \emph{et~al.}, ``Deep exploration as a unifying account of
  explore-exploit behavior,'' 2020.

\bibitem{wilson2021balancing}
R.~C. Wilson, E.~Bonawitz \emph{et~al.}, ``Balancing exploration and
  exploitation with information and randomization,'' \emph{Current opinion in
  behavioral sciences}, vol.~38, pp. 49--56, 2021.

\bibitem{tipper_object-centred_1991}
S.~P. Tipper, J.~Driver, and B.~Weaver,
  ``\BIBforeignlanguage{eng}{Object-centred inhibition of return of visual
  attention},'' \emph{\BIBforeignlanguage{eng}{The Quarterly Journal of
  Experimental Psychology. A, Human Experimental Psychology}}, vol.~43, no.~2,
  pp. 289--298, May 1991.

\bibitem{klein_inhibition_2000}
R.~M. Klein, ``\BIBforeignlanguage{en}{Inhibition of return},''
  \emph{\BIBforeignlanguage{en}{Trends in Cognitive Sciences}}, vol.~4, no.~4,
  pp. 138--147, Apr. 2000.

\bibitem{lupianez2013inhibition}
J.~Lupi{\'a}{\~n}ez, E.~Mart{\'\i}n-Ar{\'e}valo, and A.~B. Chica, ``Is
  inhibition of return due to attentional disengagement or to a detection cost?
  the detection cost theory of ior.'' \emph{Psicologica: International Journal
  of Methodology and Experimental Psychology}, vol.~34, no.~2, pp. 221--252,
  2013.

\bibitem{chica_dissociating_2006}
A.~B. Chica, J.~Lupianez, and P.~Bartolomeo,
  ``\BIBforeignlanguage{eng}{Dissociating inhibition of return from endogenous
  orienting of spatial attention: {Evidence} from detection and discrimination
  tasks},'' \emph{\BIBforeignlanguage{eng}{Cognitive Neuropsychology}},
  vol.~23, no.~7, pp. 1015--1034, Oct. 2006.

\bibitem{henderickx_involvement_2012}
D.~Henderickx, K.~Maetens, and E.~Soetens, ``\BIBforeignlanguage{en}{The
  involvement of bottom-up saliency processing in endogenous inhibition of
  return},'' \emph{\BIBforeignlanguage{en}{Attention, Perception, \&
  Psychophysics}}, vol.~74, no.~2, pp. 285--299, Feb. 2012.

\bibitem{tipper1998action}
S.~P. Tipper, L.~A. Howard, and G.~Houghton, ``Action--based mechanisms of
  attention,'' \emph{Philosophical Transactions of the Royal Society of London.
  Series B: Biological Sciences}, vol. 353, no. 1373, pp. 1385--1393, 1998.

\bibitem{howard1999inhibition}
L.~A. Howard, J.~Lupi{\'a}{\~n}ez, and S.~P. Tipper, ``Inhibition of return in
  a selective reaching task: An investigation of reference frames,'' \emph{The
  Journal of general psychology}, vol. 126, no.~4, pp. 421--442, 1999.

\bibitem{grison2004object}
S.~Grison, K.~Kessler \emph{et~al.}, ``Object-and location-based inhibition in
  goal-directed action: inhibition of return reveals behavioural and anatomical
  associations,'' in \emph{Attention in Action}.\hskip 1em plus 0.5em minus
  0.4em\relax Psychology Press, 2004, pp. 187--224.

\bibitem{sara_orienting_2012}
S.~J. Sara and S.~Bouret, ``\BIBforeignlanguage{en}{Orienting and
  {Reorienting}: {The} {Locus} {Coeruleus} {Mediates} {Cognition} through
  {Arousal}},'' \emph{\BIBforeignlanguage{en}{Neuron}}, vol.~76, no.~1, pp.
  130--141, Oct. 2012.

\bibitem{wallis_contrasting_2011}
J.~D. Wallis and S.~W. Kennerley, ``\BIBforeignlanguage{en}{Contrasting reward
  signals in the orbitofrontal cortex and anterior cingulate cortex},''
  \emph{\BIBforeignlanguage{en}{Annals of the New York Academy of Sciences}},
  vol. 1239, no.~1, pp. 33--42, 2011.

\bibitem{khani_partially_2014}
A.~Khani, ``Partially dissociable roles of {OFC} and {ACC} in stimulus-guided
  and action-guided decision making,'' \emph{Journal of Neurophysiology}, vol.
  111, no.~9, pp. 1717--1720, May 2014.

\bibitem{monosov_anterior_2017}
I.~E. Monosov, ``\BIBforeignlanguage{en}{Anterior cingulate is a source of
  valence-specific information about value and uncertainty},''
  \emph{\BIBforeignlanguage{en}{Nature Communications}}, vol.~8, no.~1, p. 134,
  Jul. 2017.

\bibitem{bar2003information}
I.~Bar-Gad, G.~Morris, and H.~Bergman, ``Information processing, dimensionality
  reduction and reinforcement learning in the basal ganglia,'' \emph{Progress
  in neurobiology}, vol.~71, no.~6, pp. 439--473, 2003.

\bibitem{santucci2010biological}
V.~G. Santucci, G.~Baldassarre, and M.~Mirolli, ``Biological cumulative
  learning through intrinsic motivations: A simulated robotic study on the
  development of visually-guided reaching.'' in \emph{EpiRob}, 2010.

\bibitem{mirolli2013phasic}
M.~Mirolli, V.~G. Santucci, and G.~Baldassarre, ``Phasic dopamine as a
  prediction error of intrinsic and extrinsic reinforcements driving both
  action acquisition and reward maximization: A simulated robotic study,''
  \emph{Neural Networks}, vol.~39, pp. 40--51, 2013.

\bibitem{ranjbar-slamloo_dopamine_2020}
Y.~Ranjbar-Slamloo and Z.~Fazlali, ``Dopamine and {Noradrenaline} in the
  {Brain}; {Overlapping} or {Dissociate} {Functions}?'' \emph{Frontiers in
  Molecular Neuroscience}, vol.~12, p. 334, 2020.

\bibitem{BOURET2005}
S.~Bouret and S.~J. Sara, ``Network reset: a simplified overarching theory of
  locus coeruleus noradrenaline function,'' \emph{Trends in Neurosciences},
  vol.~28, no.~11, pp. 574--582, 2005.

\bibitem{bouret_sensitivity_2015}
S.~Bouret and B.~J. Richmond, ``\BIBforeignlanguage{en}{Sensitivity of {Locus}
  {Ceruleus} {Neurons} to {Reward} {Value} for {Goal}-{Directed} {Actions}},''
  \emph{\BIBforeignlanguage{en}{Journal of Neuroscience}}, vol.~35, no.~9, pp.
  4005--4014, Mar. 2015.

\bibitem{jahn_noradrenergic_2020}
C.~I. Jahn, C.~Varazzani \emph{et~al.}, ``Noradrenergic {But} {Not}
  {Dopaminergic} {Neurons} {Signal} {Task} {State} {Changes} and {Predict}
  {Reengagement} {After} a {Failure},'' \emph{Cerebral Cortex (New York, NY)},
  vol.~30, no.~9, pp. 4979--4994, Jul. 2020.

\bibitem{ahmadlou_subcortical_2025}
M.~Ahmadlou, M.~Y. Shirazi \emph{et~al.}, ``A subcortical switchboard for
  perseverative, exploratory and disengaged states,'' \emph{Nature}, vol. 641,
  no. 8061, pp. 151--161, May 2025.

\bibitem{pecheux_habituation_1983}
M.-G. Pecheux and R.~Lécuyer, ``Habituation {Rate} and {Free} {Exploration}
  {Tempo} in 4-{Month}-{Old} {Infants},'' \emph{International Journal of
  Behavioral Development}, vol.~6, no.~1, pp. 37--50, Mar. 1983.

\bibitem{hunter1983effects}
M.~A. Hunter, E.~W. Ames, and R.~Koopman, ``Effects of stimulus complexity and
  familiarization time on infant preferences for novel and familiar stimuli.''
  \emph{Developmental Psychology}, vol.~19, no.~3, p. 338, 1983.

\bibitem{ruff1986components}
H.~A. Ruff, ``Components of attention during infants' manipulative
  exploration,'' \emph{Child development}, pp. 105--114, 1986.

\bibitem{schmid_habituation_2015}
S.~Schmid, D.~A. Wilson, and C.~H. Rankin, ``Habituation mechanisms and their
  importance for cognitive function,'' \emph{Frontiers in Integrative
  Neuroscience}, vol.~8, p.~97, Jan. 2015.

\bibitem{schoner2006using}
G.~Sch{\"o}ner and E.~Thelen, ``Using dynamic field theory to rethink infant
  habituation.'' \emph{Psychological review}, vol. 113, no.~2, p. 273, 2006.

\bibitem{perone2013autonomy}
S.~Perone and J.~P. Spencer, ``Autonomy in action: linking the act of looking
  to memory formation in infancy via dynamic neural fields,'' \emph{Cognitive
  science}, vol.~37, no.~1, pp. 1--60, 2013.

\bibitem{yarrow_infants_1982}
L.~J. Yarrow, G.~A. Morgan \emph{et~al.}, ``\BIBforeignlanguage{en}{Infants'
  persistence at tasks: {Relationships} to cognitive functioning and early
  experience},'' \emph{\BIBforeignlanguage{en}{Infant Behavior and
  Development}}, vol.~5, no.~2, pp. 131--141, Jan. 1982.

\bibitem{hupp1991persistence}
S.~C. Hupp and L.~Abbeduto, ``Persistence as an indicator of mastery motivation
  in young children with cognitive delays,'' \emph{Journal of Early
  Intervention}, vol.~15, no.~3, pp. 219--225, 1991.

\bibitem{teubner-rhodes_cognitive_2017}
S.~Teubner-Rhodes, K.~I. Vaden \emph{et~al.},
  ``\BIBforeignlanguage{en}{Cognitive persistence: {Development} and validation
  of a novel measure from the {Wisconsin} {Card} {Sorting} {Test}},''
  \emph{\BIBforeignlanguage{en}{Neuropsychologia}}, vol. 102, pp. 95--108, Jul.
  2017.

\bibitem{teubner-rhodes_cognitive_2020}
S.~Teubner-Rhodes, ``Cognitive {Persistence} and {Executive} {Function} in the
  {Multilingual} {Brain} {During} {Aging},'' \emph{Frontiers in Psychology},
  vol.~11, 2020.

\bibitem{matias2017activity}
S.~Matias, E.~Lottem \emph{et~al.}, ``Activity patterns of serotonin neurons
  underlying cognitive flexibility,'' \emph{Elife}, vol.~6, 2017.

\bibitem{lottem2018activation}
E.~Lottem, D.~Banerjee \emph{et~al.}, ``Activation of serotonin neurons
  promotes active persistence in a probabilistic foraging task,'' \emph{Nature
  communications}, vol.~9, no.~1, p. 1000, 2018.

\bibitem{BussMagnottaPennyEtAl2021}
A.~T. Buss, V.~A. Magnotta \emph{et~al.}, ``How do neural processes give rise
  to cognition? simultaneously predicting brain and behavior with a dynamic
  model of visual working memory,'' \emph{Psychological review}, vol. 128,
  no.~2, p. 362–395, 2021.

\bibitem{TekulveSchoner2020}
J.~Tek{\"u}lve and G.~Sch{\"o}ner, ``A neural dynamic network drives an
  intentional agent that autonomously learns beliefs in continuous time,''
  \emph{IEEE Transactions on Cognitive and Developmental Systems}, p. 1–12,
  2020.

\bibitem{AerdkerFengSchoner2022}
S.~Aerdker, J.~Feng, and G.~Sch{\"o}ner, ``Habituation and dishabituation in
  motor behavior: Experiment and neural dynamic model,'' \emph{Frontiers in
  Psychology}, vol.~13, p. 717669, April 2022.

\bibitem{ijspeert_dynamical_2013}
A.~J. Ijspeert, J.~Nakanishi \emph{et~al.}, ``Dynamical {Movement}
  {Primitives}: {Learning} {Attractor} {Models} for {Motor} {Behaviors},''
  \emph{Neural Computation}, vol.~25, no.~2, pp. 328--373, Feb. 2013.

\bibitem{pastor2013dynamic}
P.~Pastor, M.~Kalakrishnan \emph{et~al.}, ``From dynamic movement primitives to
  associative skill memories,'' \emph{Robotics and Autonomous Systems},
  vol.~61, no.~4, pp. 351--361, 2013.

\bibitem{pastor2009learning}
P.~Pastor, H.~Hoffmann \emph{et~al.}, ``Learning and generalization of motor
  skills by learning from demonstration,'' in \emph{2009 IEEE International
  Conference on Robotics and Automation}.\hskip 1em plus 0.5em minus
  0.4em\relax IEEE, 2009, pp. 763--768.

\bibitem{lomp2013software}
O.~Lomp, S.~K.~U. Zibner \emph{et~al.}, ``A software framework for cognition,
  embodiment, dynamics, and autonomy in robotics: cedar,'' in \emph{Artificial
  Neural Networks and Machine Learning--ICANN 2013: 23rd International
  Conference on Artificial Neural Networks Sofia, Bulgaria, September 10-13,
  2013. Proceedings 23}.\hskip 1em plus 0.5em minus 0.4em\relax Springer, 2013,
  pp. 475--482.

\bibitem{ros}
\BIBentryALTinterwordspacing
{Stanford Artificial Intelligence Laboratory et al.}, ``Robotic operating
  system.'' [Online]. Available: \url{https://www.ros.org}
\BIBentrySTDinterwordspacing

\bibitem{houbreinh}
Q.~Houbre, A.~Angleraud, and R.~Pieters, ``An inhibition of return mechanism
  for the exploration of sensorimotor contingencies,'' in \emph{2020 IEEE
  International Conference on Human-Machine Systems (ICHMS)}, 2020, pp. 1--6.

\bibitem{houbreexpl}
Q.~Houbre and R.~Pieters, ``Active exploration and working memory synaptic
  plasticity shapes goal-directed behavior in curiosity-driven learning,''
  \emph{Cognitive Systems Research}, vol.~91, p. 101339, 2025.

\bibitem{amari_dynamics_1977}
S.-i. Amari, ``\BIBforeignlanguage{en}{Dynamics of pattern formation in
  lateral-inhibition type neural fields},''
  \emph{\BIBforeignlanguage{en}{Biological Cybernetics}}, vol.~27, no.~2, pp.
  77--87, Jun. 1977.

\bibitem{mcclelland_why_1995}
J.~L. McClelland, B.~L. McNaughton, and R.~C. O'Reilly,
  ``\BIBforeignlanguage{eng}{Why there are complementary learning systems in
  the hippocampus and neocortex: insights from the successes and failures of
  connectionist models of learning and memory},''
  \emph{\BIBforeignlanguage{eng}{Psychological Review}}, vol. 102, no.~3, pp.
  419--457, Jul. 1995.

\bibitem{parisi2019continual}
G.~I. Parisi, R.~Kemker \emph{et~al.}, ``Continual lifelong learning with
  neural networks: A review,'' \emph{Neural networks}, vol. 113, pp. 54--71,
  2019.

\end{thebibliography}
\newpage
\appendix

\subsection{Dynamic Field Theory}\label{app:dft}

Dynamic neural fields (DNF) represent the distribution of neural populations and their evolution in time according to Amari's equation \cite{amari_dynamics_1977} :
\begin{equation}
\begin{split}
    \tau\dot{u}(x,t) = -u(x,t) &+ h + S(x,t) + \xi(x,t) \\
    &+ \int \sigma(u(x',t)) \omega(x - x') dx' 
\end{split}
\end{equation}

\begin{figure}[h]
\centering
\includegraphics[width=0.45\textwidth]{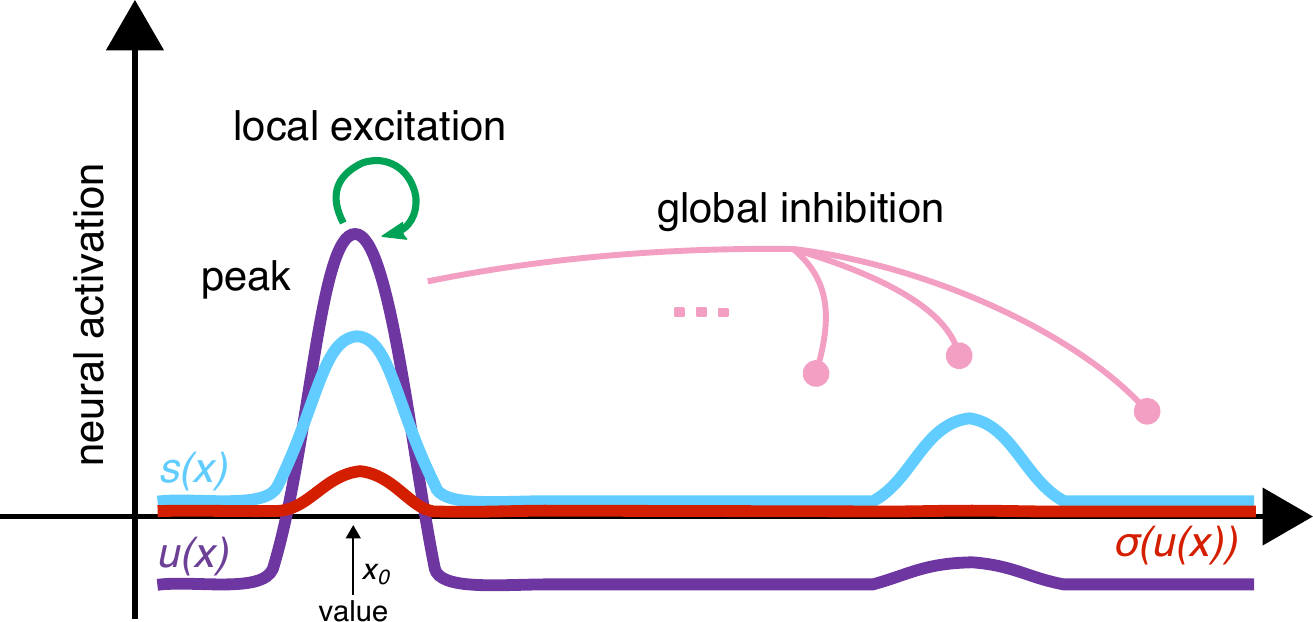}
\caption{One dimensional neural field\label{fig:amari}}
\end{figure}

with \textit{u(x,t)} the activation field over a feature dimension \textit{x} at time \textit{t}. The resting level is \textit{h} (\textit{h $<$ 0}), \textit{S(x,t)} is the external input and $\tau$ is a time constant. Furthermore, we define \textit{$\xi$(x,t)} as noise, which is useful to generate a stochastic peak of activation if the resting level is sufficiently high. An output signal \textit{$\sigma$(u(x,t))} is determined by activation \textit{ u (x, t)} through a sigmoid function with a threshold set to zero. An interaction kernel $\omega$ is used as a convolution with the output \textit{$\sigma$(u(x,t))} and serves as local excitation and surrounding inhibition. Figure \ref{fig:amari} depicts a one-dimensional field according to Amari's equation. The influence of the Gaussian kernel is significant, as varying shapes can impact the neural dynamics of a field. For instance, local excitatory (bell-shaped) coupling stabilizes peaks against decay, while lateral inhibitory coupling (Mexican hat shape) prevents the activation from spreading out along the neural field. Based on the interaction kernel, a neural field can operate in several modes. In a self-stabilized mode, peaks of activation are stabilized against input noise. In a self-sustained mode, the field exhibits suprathreshold peaks even in the absence of external activation. This mode allows us to model working memory fields (WM) in our approach. A selective mode is also possible through lateral inhibition that allows the emergence of a single peak of activation.

In addition to the neural field, DFT also defines a memory trace :
\begin{equation}
\begin{split}
    \dot{v}(t) = a(t) (\frac{1}{\tau_+}& (-v(t)+f(u(t)))f(u(t)) \\
    &+ \frac{1}{\tau_-}(-v(t)(1-f(u(t)))) , 
\end{split}
\label{eq:mt}
\end{equation}
A memory trace basically builds an activation depending on the time constant \textit{$\tau_+$} and this activation decays according to \textit{$\tau_-$}. In this specific memory trace, we introduce \textit{a(t)} as an activation coming from a zero-dimensional field. With this term, the dynamic of the memory trace can take place only if it receives an additional activation from a node. This specific field is useful for retaining various activations of different intensities in time and will be the core element to compute the prediction error and the learning progress while the robot is pursuing a goal. 

We define a slow boost component, a memory trace that takes two inputs. When the active node propagates excitation, activation increases; however, it remains steady if both linked nodes are low and decreases if only $\sigma(c_{thr})$ is high.
The equation is as follows :
\begin{equation}
\begin{split}
    \dot{v}(t) = \sigma(c_{act}(t)) & \biggr[ \frac{1}{\tau_+} (-v(t)+\sigma(c_{act}(t)))\sigma(c_{act}(t)) \biggr] \\
    &+ \sigma(c_{thr}(t)) \biggr[ \frac{1}{\tau_-}(-v(t)(1-\sigma(c_{act}(t))) \biggr] , 
\end{split}
\label{eq:slowboost}
\end{equation}
with \textit{$\sigma(c_{act})$} the absolute sigmoid activation coming from the active node and \textit{$\sigma(c_{thr})$} the same activation function coming from the threshold node.

We apply a convolution on the output of the object position within the action formation module. This convolution takes the shape of a Mexican hat, and the result of this convolution is used as an inhibition to the action formation NF. A Mexican hat shape can be described as the difference of two Gaussians with a narrow excitatory component and a wider inhibitory component. In our case, the convolution takes the form :
\begin{equation}
\begin{split}
    k(x,y) = c_{exc}.exp\left[-\frac{1}{2} \left( \frac{x^2}{\sigma_{x,exc}^2}+\frac{y^2}{\sigma_{y,exc}^2}\right) \right] \\ 
    - c_{inh}.exp\left[-\frac{1}{2} \left( \frac{x^2}{\sigma_{x,inh}^2}+\frac{y^2}{\sigma_{y,inh}^2}\right) \right]
\end{split}
\end{equation}
with $c_{exc}$ the strength of the lateral excitation, $\sigma_{x,exc}$ and $\sigma_{y,exc}$ the standard deviation along each dimension. $c_{inh}$, $\sigma_{x,inh}$ and $\sigma_{y,inh}$ are the parameters for Gaussian inhibition. In our experiment, the parameters are: $c_{exc}$ = 3.5, $\sigma_{x,exc}$ = 2.0, $\sigma_{y,exc}$ = 2.0, $c_{inh}$ = -4.0, $\sigma_{x,inh}$ = 9.0 and $\sigma_{y,inh}$ = 9.0.

The input from position object NF with the convolution is the following :
\begin{equation}
    s(x,y) = - (a_s . k(x,y))
\end{equation}

with $a_s$ the stimulus strength of the position of the object (1.0).
\subsection{Dynamic Motion Primitives}\label{app:dmp}

\begin{figure*}
\centering
\subfloat{\includegraphics[width=2.2in]{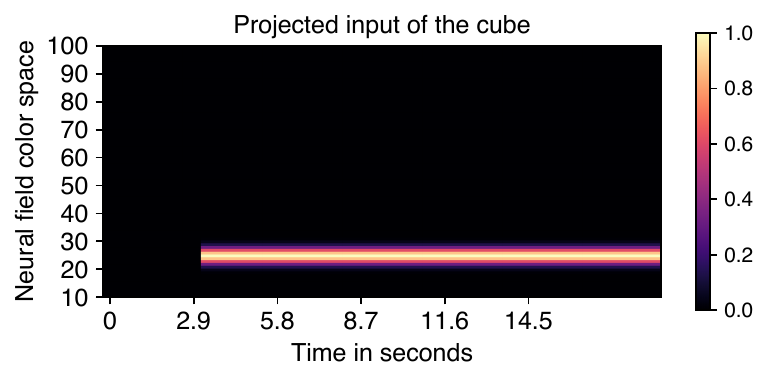}}
\hfil
\subfloat{\includegraphics[width=2.4in]{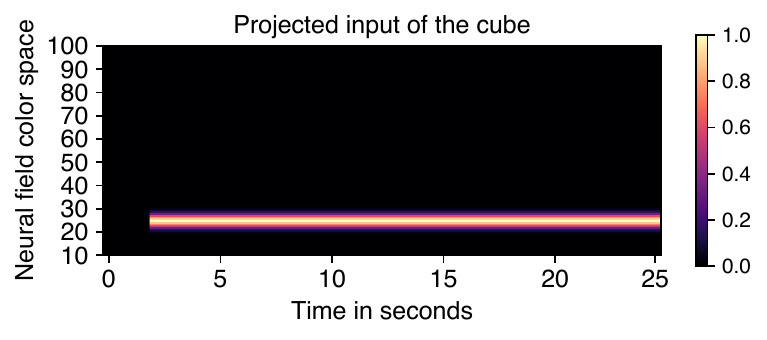}}
\hfil
\subfloat{\includegraphics[width=2.4in]{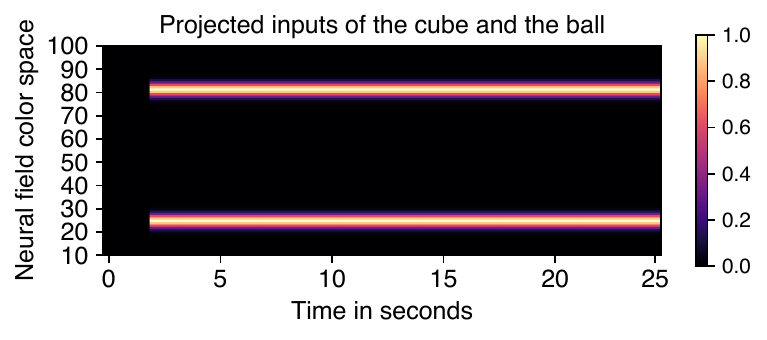}}
\hfil
\subfloat{\includegraphics[width=2.2in]{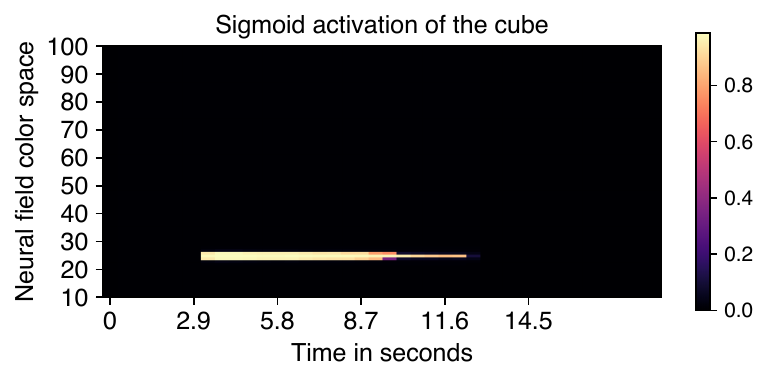}}
\hfil
\subfloat{\includegraphics[width=2.4in]{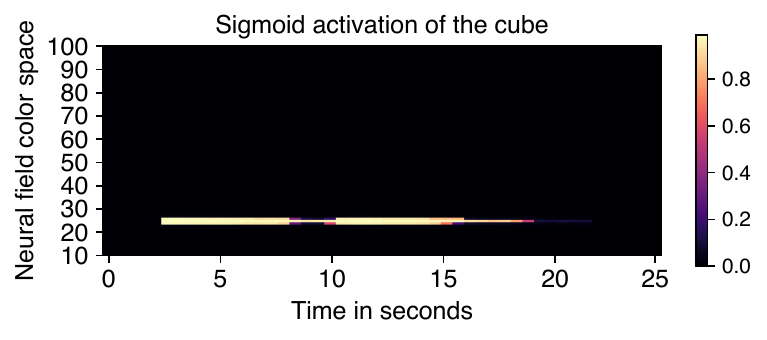}}
\hfil
\subfloat{\includegraphics[width=2.4in]{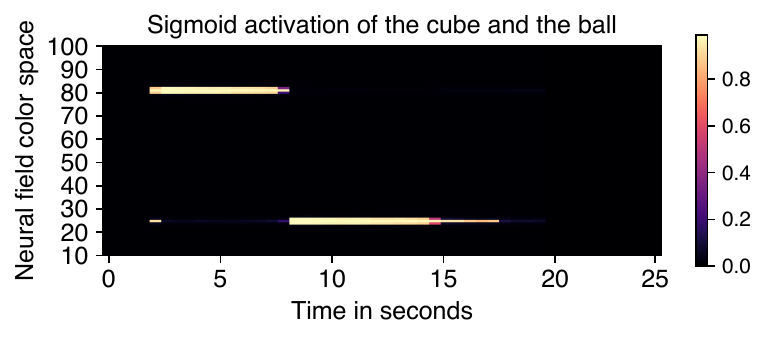}}
\hfil
\subfloat{\includegraphics[width=2.3in]{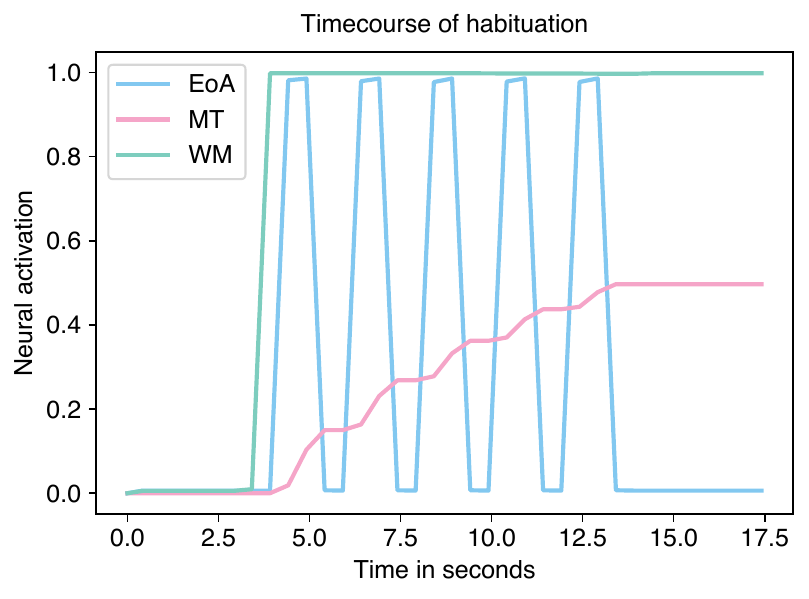}}
\hfil
\subfloat{\includegraphics[width=2.3in]{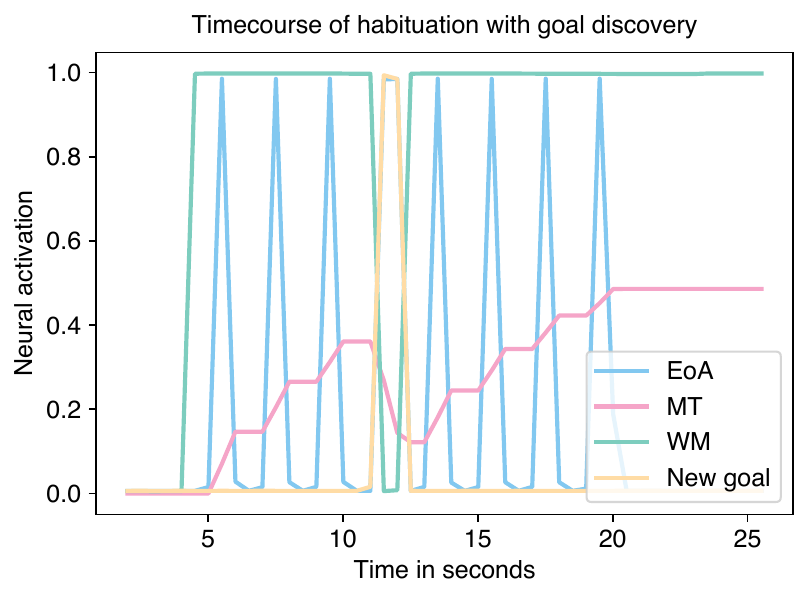}}
\hfil
\subfloat{\includegraphics[width=2.3in]{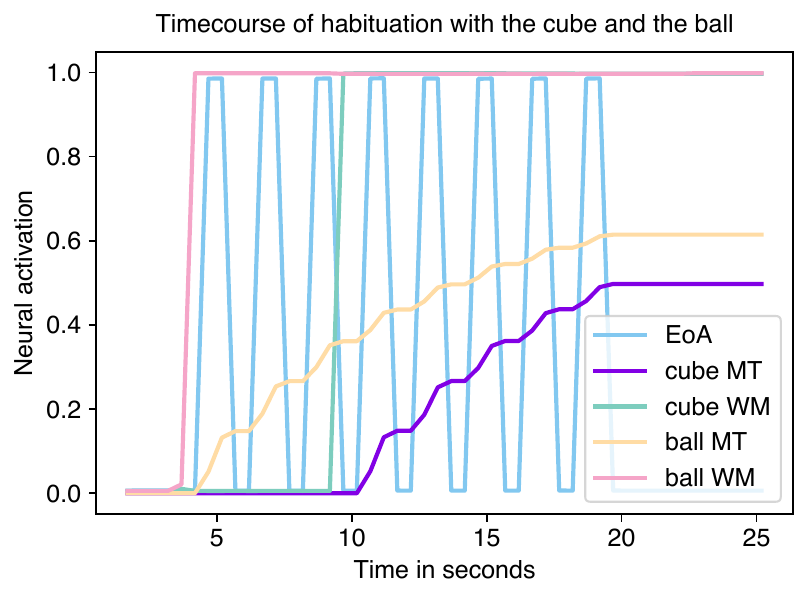}}%
\caption{Simulation of the habituation module under three conditions. The left column depicts the cube habituation with no new goal discovery. The middle column simulates a cube habituation with the discovery of a new goal. The right column introduces the cube and ball habituation with no goal discovery. The first row represents the input projected within the object selection NF. The second row captures the sigmoid activation within the object selection NF. The third row represents the End of Action node (EoA), the working memory (WM), and memory trace (MT) activity within CoS EOA, WM colors, and visual memory, respectively.}
\label{fig:sim_habit}
\end{figure*}

Originally, the method for generating a one-dimensional primitive employs a set of differential equations \cite{ijspeert_dynamical_2013}. Here, we use a derived version \cite{pastor2009learning} :
\begin{equation}
    \tau \dot{v} = K(g - x) - Dv - K(g - x_0)s + Kf(s)
\label{eq:dyn}
\end{equation}
\begin{equation}
    \tau \dot{x} = v 
\end{equation}
with $x$ and $v$ are the position and velocity, $x_0$ and $g$ are the starting position and the target position, and $\tau$ is a temporal scaling factor. $K$ serves as a spring constant, the system is critically damped with $D$, and $f$ is a nonlinear function defined by :
\begin{equation}
    f(s) = \frac{\Sigma_i \omega_i \psi_i(s)s}{\Sigma_i \psi_i(s)s}
\label{eq:dyn2}
\end{equation}
\begin{equation}
    \psi_i(s) = \exp(-h_i(s - c_i)^2)
\label{eq:gaussdyn}
\end{equation}
Equation (\ref{eq:gaussdyn}) represents Gaussian basis functions with center $c_i$, width $h_i,$ and adjustable weights $\omega_i$. 
The function $f$ depends on a phase variable $s$, which monotonically changes from 1 to 0 during a movement, with the equation :
\begin{equation}
    \tau \dot{s} = -\alpha s
    \label{eq:monodyn}
\end{equation}
with $\alpha$ as a predefined constant.
To learn a motion from a demonstration, a movement $x(t)$ is recorded, and we compute its derivatives $v(t)$ and $\dot{v}(t)$ for each time step $t = [0...T]$. Then $s(t)$ is computed according to a relevant temporal scale $\tau$. After this, we can compute $f_{target}(s)$ for each values:
\begin{equation}
    f_{target}(s) = \frac{\tau \dot{v} + Dv}{K} - (g - x) + (g - x_0)s
\end{equation}
where $x_0$ and $g$ are set to $x(0)$ and $x(T)$. Then, computing the weights in equation (\ref{eq:dyn2}) that minimize the error criterion can be done with linear regression: $J = \Sigma_s(f_{target}(s) - f(s))^2$.
To generate a new motion, the weights $\omega_i$ are reused by specifying a start point ($x_0$), a stop point ($g$), and setting $s$ to 1. 

In our experiment, we record the movement of the end effector at 20Hz, gathering about 40 different 3D points for a trajectory. If the motion results in the discovery of a goal, the trajectory points are used to generate the DMPs (one per dimension). To generate a movement with the DMPs parameters, we only have to specify a target point $g$. In fact, we keep the same initial position $x_0$ that corresponds to the resting state of the robot.

The following parameters are used to generate a new trajectory after learning: $K = 100$, $D = 20$, $\tau = 2 * \tau_{demo}$ with $\tau_{demo}$ being the length of the demonstration. The learning and generation of DMPs is performed through the dmp ROS package\footnote{https://wiki.ros.org/dmp}.

\begin{figure*}
\centering
\subfloat{\includegraphics[width=3.0in]{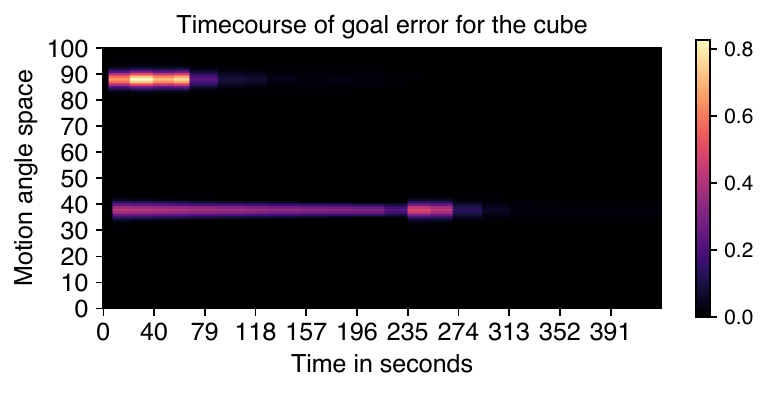}}
\hfil
\subfloat{\includegraphics[width=3.0in]{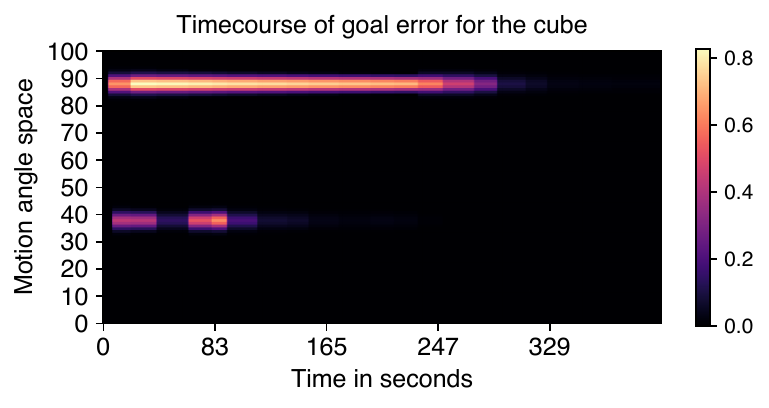}}
\hfil
\subfloat{\includegraphics[width=3.0in]{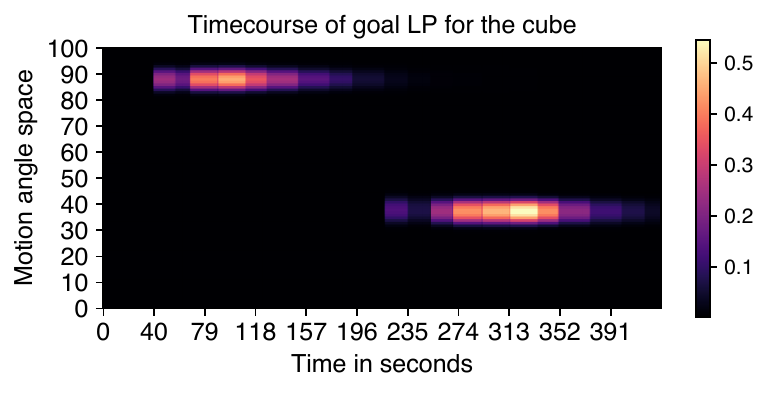}}
\hfil
\subfloat{\includegraphics[width=3.0in]{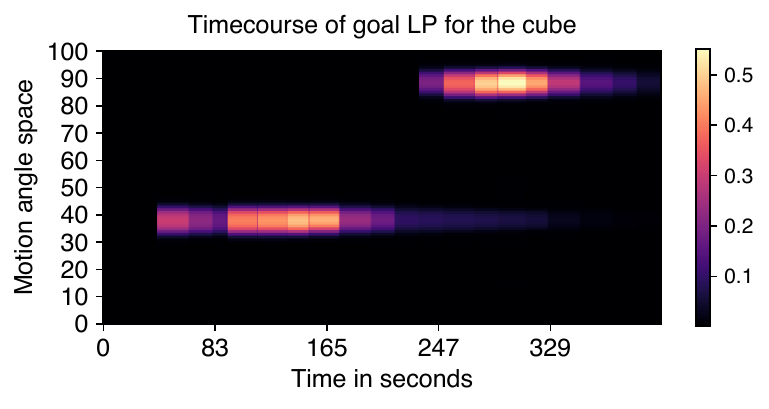}}
\hfil
\subfloat{\includegraphics[width=3.0in]{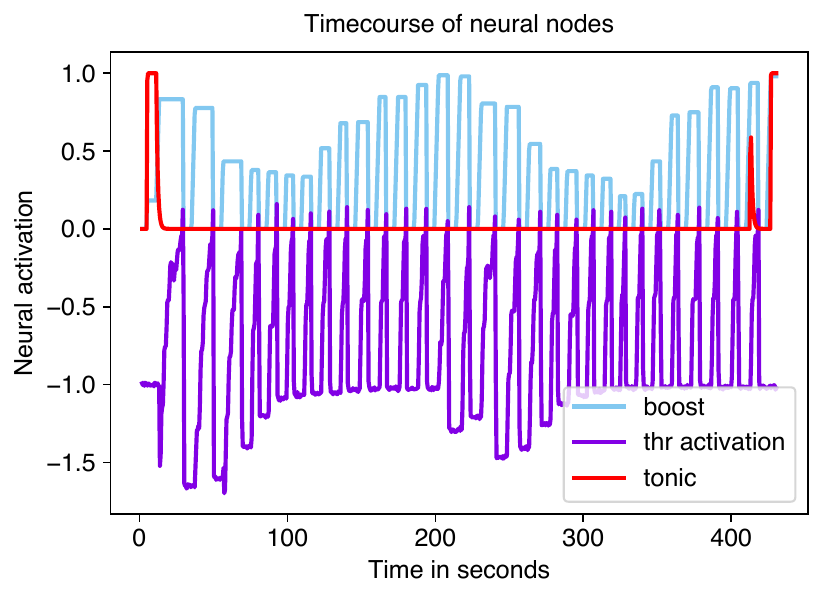}}
\hfil
\subfloat{\includegraphics[width=3.0in]{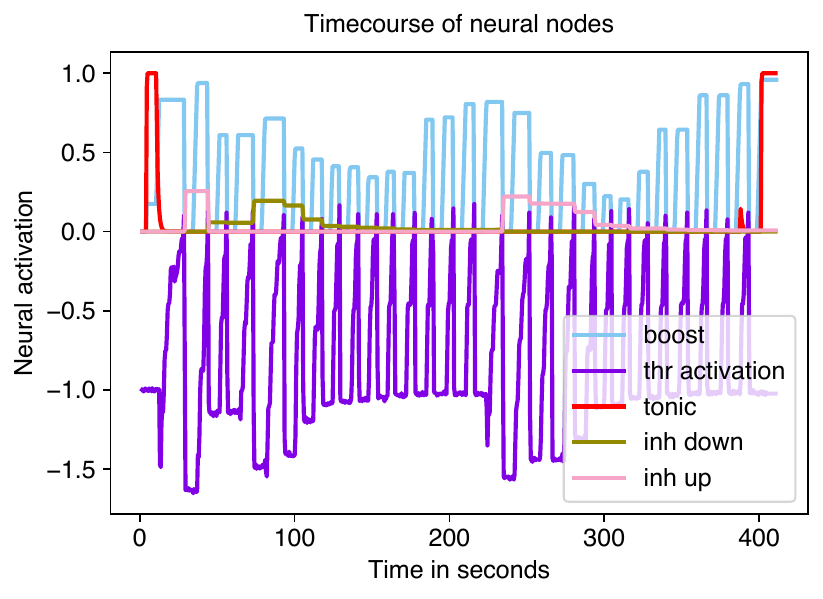}}
\caption{Simulation of the Persistence mechanism. The left column introduces a short persistence (transient action MT node $\tau_+ = 3000ms$, $\tau_- = 100ms$). The right column has the same persistence but includes the error inhibition (attempts memory trace $\tau_+ = 2000ms$ and $\tau_- = 1500ms$). The goals push-up and pull-down are respectively shown at locations 87 and 37 on the feature space.}
\label{fig:sim_persist}
\end{figure*}

\subsection{Simulation of Habituation}\label{app:habituation}

As described in section \ref{sec:habit}, the habituation paradigm is adapted from \cite{perone2013autonomy} to avoid the dehabituation of an object if this one disappears from the visual scene. We perform simulations of the habituation model under three different conditions. In the first column, we introduce the cube and let the robot explore until the object is habituated. To do so, the robot performed actions that did not lead to the discovery of new stimuli. This results in the memory trace building up activation until the inhibition is strong enough to suppress any suprathreshold activation within the object selection field. 
The second column also introduces a cube, but the robot discovered a new goal, leading to a complete drop in neural activation within the working memory and a more moderate drop in the level of activation of the memory trace.
In the third column, we simultaneously introduced a cube and a ball to the robot. The object selection NF sees a stable suprathreshold activation at the ball location. The robot then performs different actions during exploration to discover new stimuli. Without any new goal discovery, the ball MT progressively inhibits the ball peak until the object selection NF sees a peak of activation at the cube location. Finally, the robot explores the location of the cube until it becomes habituated to that object. The LC mechanism will also see the emergence of a peak in tonic NF after habituation and when all goals are learned; however, it is not possible to know in advance which object will be selected when the slow boost component raises the resting level.

\subsection{Persistence in the Locus Coeruleus}\label{app:persistence}

We evaluated the persistence mechanism with a cube in Figure \ref{fig:sim_persist}. To do so, we selected the goals \textit{push up} and \textit{pull down}, then replayed samples from the respective forward models in the experiment (Section \ref{res}). Once a goal is selected, the robot performs four trials without success before receiving a sample. The purpose of this simulation is to observe the evolution of various neural dynamics in the case of a short persistence with no error inhibition (left column in figure \ref{fig:sim_persist}) and for a short persistence with error inhibition (right column). In the first case, the goal with the highest error is selected, and a rise in learning progress (LP) is quickly seen. At first, the threshold activation is significantly lower when the error remains high. This is due to the inhibition received from the goal error MT. Without inhibition, this node has a resting level of -1 and would need 3 trials before excitation from the transient action node causes a suprathreshold peak of activation. Then, a significant error will increase the number of attempts to learn a goal.

\begin{figure}[h]
\centering
\subfloat{\includegraphics[width=3.2in]{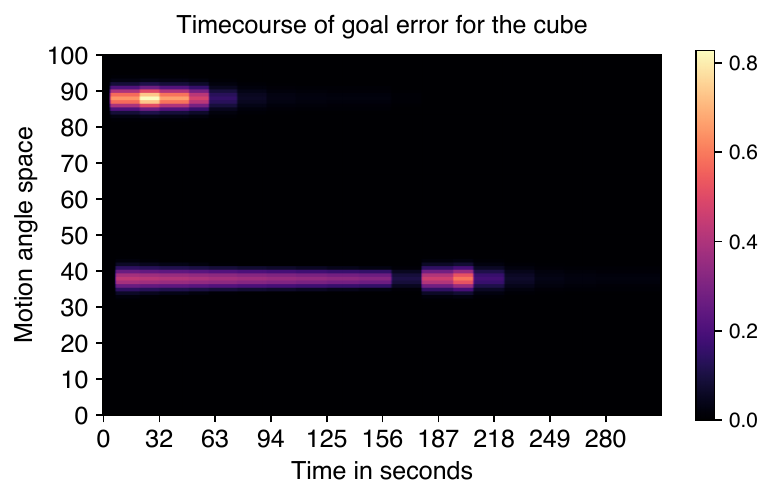}}
\hfil
\subfloat{\includegraphics[width=3.2in]{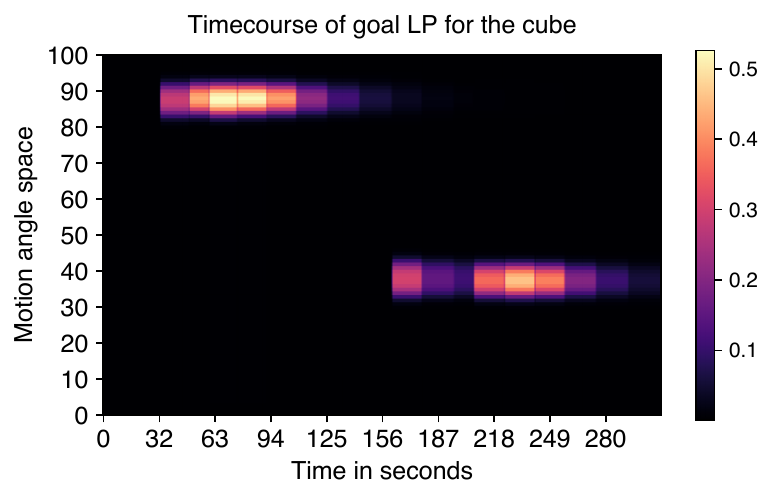}}
\hfil
\subfloat{\includegraphics[width=3.2in]{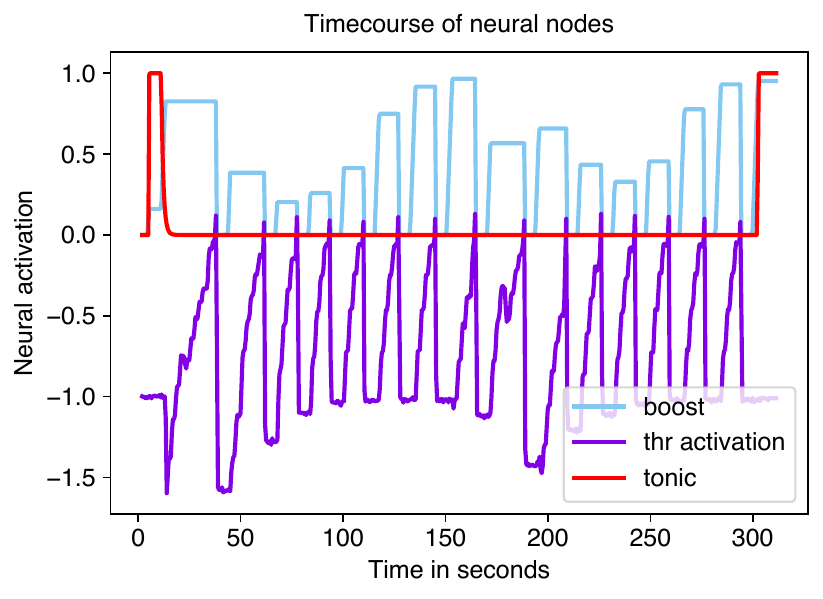}}
\caption{Simulation of the persistence module with a long persistence (transient action MT node $\tau_+ = 6000ms$, $\tau_- = 100ms$). The goals push-up and pull-down are respectively shown at locations 87 and 37 on the feature space. The bottom row represents the activation within the boost node, as well as the tonic field within LC, and the threshold activation inside persistence. At 180s, a rise in the pull-down error (top row) directly impacts the threshold activation.}
\label{fig:sim_long_persist}
\end{figure}

The second condition introduces inhibition from the attempts memory trace. Its role is to avoid selecting a goal whose error remains high without generating learning progress. In this case, the \textit{push up} goal is selected first for learning, but fails to lower the error. At time $\approx$ 40s, the \textit{pull down} goal is selected and creates a rise in learning progress, which will determine the robot's engagement in learning that goal. When the first goal has been mastered, the robot can continue learning \textit{push up}.

In figure \ref{fig:sim_long_persist}, we simulate long persistence and observe the learning of the goals \textit{push up} and \textit{pull down}. By increasing the $\tau_+$ value of the transient action MT node, the number of attempts naturally increases. It provides the robot with time to see the emergence of learning progress for both goals. We reproduced the same simulation by including the error inhibition and obtained the same results. However, this does not mean that the error inhibition is useless in the case of long persistence. If the exploration space is large and the amount of successful action is more sparse, error inhibition will remain useful for switching goals if a goal is too difficult to learn.

\subsection{Perception}\label{app:perception}

\begin{figure}[ht]
\centering
\subfloat{ 
%\hspace{2mm}
\includegraphics[width=1.3in]{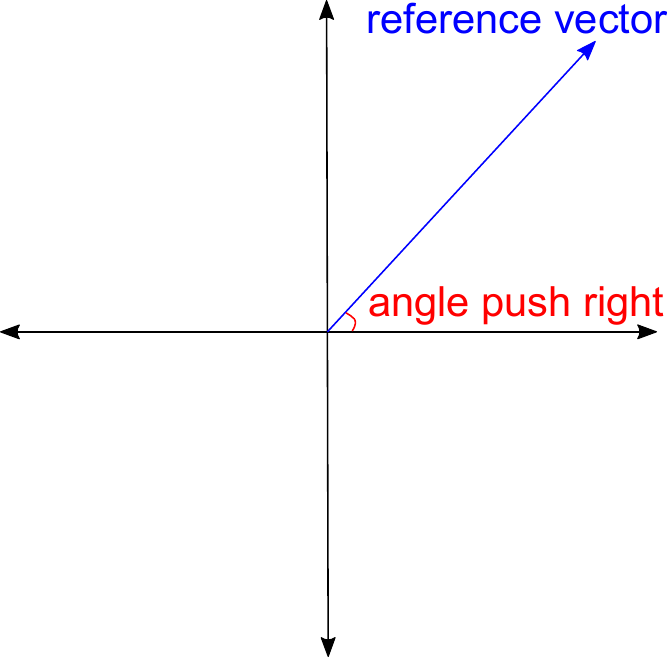}}
\hfil
\subfloat{
%\hspace{0mm}6
\includegraphics[width=1.9in]{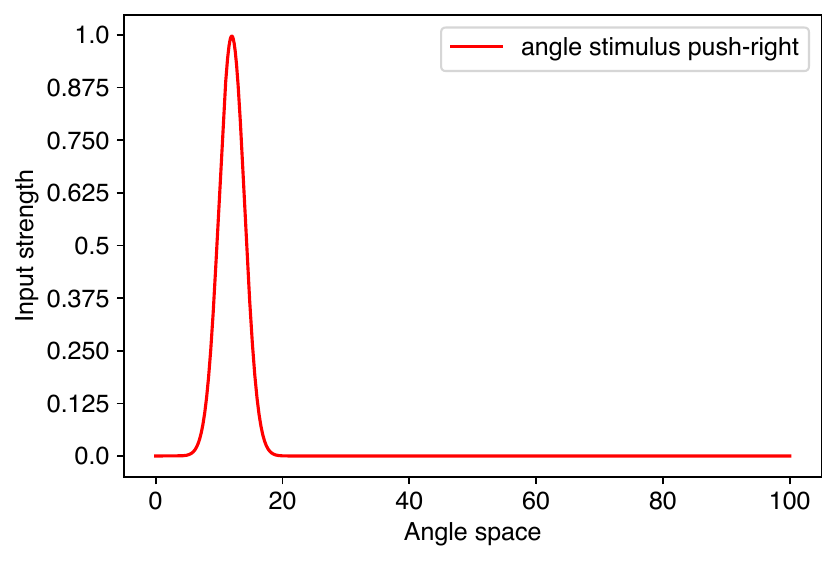}}
\caption{(Left) Angle of the object being pushed to the right. (Right) This angle is scaled and normalized as a Gaussian activation. \label{fig:angle}}
\end{figure}

The goal direction of an object is formalized by an angle between a reference vector and the vector of the motion of the object (Figure \ref{fig:angle}).
The angle is measured in radians between $-\pi$ and $\pi$ and then scaled between \textit{0} and \textit{100} to fit the dynamic neural field feature space. The objects are represented by their color within the habituation architecture. The color spectrum goes from red to blue and fits the neural field feature space from 0 to 100.

\subsection{Neural networks training}\label{app:training}

The synaptic weights of each multilayer perceptron are initialized with the same parameters to avoid any bias between objects and goals. We endow both forward and inverse models with a buffer of 20 samples to avoid catastrophic forgetting \cite{mcclelland_why_1995}, \cite{parisi2019continual}. The output error of the neural networks is calculated with the mean square error (MSE). To avoid receiving a loss superior to one, the error is combined with the hyperbolic tangent function. In this manner, we potentially avoid having multiple goals reaching a peak level superior to one within the neural fields (i.e., error module). When the models receive new samples, the training performs only one backpropagation with them. Then, the samples are added to the memory of their respective model (i.e.,forward and inverse), and the training continues for one epoch on the entire buffer memory. 

\subsection{Open access}

The source code to reproduce the experiment can be found in \textit{Anonymous submission}. The architecture file simulating the dynamic neural fields with Cedar, as well as a demo of the experiment are available.

\end{document}